\documentclass[10pt,journal,compsoc]{IEEEtran}
\usepackage{amsmath,amssymb,amsfonts}
\usepackage{algorithmic}
\usepackage{graphicx}
\usepackage{textcomp}
\usepackage{xcolor}
\usepackage{booktabs} % For formal tables
\usepackage{algorithm}
\usepackage{multirow}
\usepackage{comment}
\usepackage{subfig}
\usepackage{comment}
\usepackage{array,multirow}
\usepackage{epstopdf}
\usepackage{xspace}

\newtheorem{infproblem}{Informal Problem}
\newtheorem{problem}{Problem}
\newtheorem{theorem}{Theorem}
\newtheorem{proposition}{Proposition}

\newcommand{\tensorglue}{\textsc{Prema}\xspace}
\newcommand{\blind}{\textsc{B-Prema}\xspace}

\newcommand{\R}{\mathbb{R}}

\newcommand{\tensorX}{\underline{{\mathbf{X}}}}

\newcommand{\tensorY}{\underline{{\mathbf{Y}}}}

\newcommand{\tensorYt}{\underline{{\mathbf{Y}}}^t}

\newcommand{\tensorYc}{\underline{{\mathbf{Y}}}^c}

\newcommand{\tensorW}{\underline{{\mathbf{\Omega}}}}
\newcommand{\tensorWt}{\underline{{\mathbf{\Omega}}}^t}
\newcommand{\tensorWc}{\underline{{\mathbf{\Omega}}}^c}

\DeclareMathOperator*{\argmin}{argmin}
\DeclareMathOperator*{\minimize}{minimize}

\usepackage{footmisc}

\ifCLASSOPTIONcompsoc
  \usepackage[nocompress]{cite}
\else
  \usepackage{cite}
\fi
\ifCLASSINFOpdf
\else
\fi
\begin{document}
\title{\tensorglue: Principled Tensor Data Recovery from Multiple Aggregated Views}
\author{Faisal~M.~Almutairi,
	Charilaos~I.~Kanatsoulis,
	and~Nicholas~D.~Sidiropoulos% <-this % stops a space
	\IEEEcompsocitemizethanks{\IEEEcompsocthanksitem F. M. Almutairi and C. I. Kanatsoulis are  with the Department of Electrical and Computer Engineering, University of Minnesota, Minneapolis, MN 55455 USA.\protect\\
		E-mails: (almut012, kanat003)@umn.edu.
		\IEEEcompsocthanksitem N. D. Sidiropoulos is with the Department of Electrical and Computer Engineering, University of Virginia, Charlottesville, VA 22903, USA.\protect\\
		E-mail: nikos@virginia.edu}
	%}
%\author{Michael~Shell,~\IEEEmembership{Member,~IEEE,}
 %       John~Doe,~\IEEEmembership{Fellow,~OSA,}
 %       and~Jane~Doe,~\IEEEmembership{Life~Fellow,~IEEE}% <-this % stops a space
%\IEEEcompsocitemizethanks{\IEEEcompsocthanksitem M. Shell was with the Department
%of Electrical and Computer Engineering, Georgia Institute of Technology, Atlanta,
%GA, 30332.\protect\\
%E-mail: see http://www.michaelshell.org/contact.html
%\IEEEcompsocthanksitem J. Doe and J. Doe are with Anonymous University.}% <-this % stops an unwanted space
\thanks{%Manuscript received April 19, 2005; revised August 26, 2015.
The work of C. I. Kanatsoulis and  N. D. Sidiropoulos was partially supported by the National Science Foundation under Grants NSF IIS-1704074, and NSF ECCS-1608961.}
}


\markboth{\tensorglue: Principled Tensor Data Recovery from Multiple Aggregated Views}%
{Almutairi \MakeLowercase{\textit{et al.}}:\tensorglue: Principled Tensor Data Recovery from Multiple Aggregated Views}
\IEEEtitleabstractindextext{%
\begin{abstract}
Multidimensional data have become ubiquitous and are frequently encountered in situations where the information is aggregated over multiple data atoms. The aggregation can be over time or other features, such as geographical location. We often have access to multiple aggregated views of the same data, each aggregated in one or more dimensions, especially when data are collected or measured by different agencies. For instance, item sales can be aggregated temporally, and over groups of stores based on their location or affiliation. However, data mining and machine learning models benefit from detailed data for personalized analysis and prediction. 
Thus, data disaggregation algorithms are becoming increasingly important in various domains. The goal of this paper is to reconstruct finer-scale data from multiple coarse views, aggregated over different (subsets of) dimensions. The proposed method, called \tensorglue, leverages low-rank tensor factorization tools to fuse the multiple views and provide recovery guarantees under certain conditions. 
\tensorglue can tackle challenging scenarios, such as missing or partially observed data, double aggregation, and even blind disaggregation (without knowledge of the aggregation patterns) using a variant of \tensorglue called \blind. To showcase the effectiveness of \tensorglue, the paper includes extensive experiments using real data from different domains: retail sales, crime counts, and  weather observations.  
\end{abstract}


}
%\end{comment}

\maketitle

\IEEEdisplaynontitleabstractindextext

% \fi
\IEEEpeerreviewmaketitle


%\vspace{-3mm}

%\begin{comment}

\IEEEraisesectionheading{\section{Introduction}\label{sec:introduction}}
%\vspace{-1mm}

\IEEEPARstart{D}{ata} aggregation is the process of summing (or averaging) multiple data samples from a certain dataset, which results in data resolution reduction and compression.   
The most common type of aggregation is {\em temporal aggregation}. For example, the annual income is the aggregate of the monthly salary. Aggregation over other attributes is also common, e.g., data get aggregated geographically (e.g., population of New York) or according to a defined affiliation (e.g., employees of Company X).  
The latter is known in economics as {\em contemporaneous aggregation} \cite{silvestrini2008temporal}. The different types of aggregation are often combined, e.g., the number of foreigners who visited different US states in $2019$ can be aggregated in time, location (states), and affiliation (nationality). 
%%%%%%%%%%%%%%%%%%%%%%%%%

In some cases, it is the data collection process that limits data resolution in the first place, e.g., Store X records item sales only on a monthly basis. 
Aggregated data also exist for other reasons, the most important being data summarization.
In particular, aggregated data enjoy concise representations, which is critical in the era of data deluge.
Aggregation also benefits various other purposes, including scalability~\cite{uludag2007analysis}, communication and storage costs~\cite{patil2010data}, and privacy~\cite{shi2011privacy}.
Aggregated data are common in a wide range of domains, such as economics, health care~\cite{park2014ludia}, education~\cite{ellaway2014developing}, wireless communication, signal and image processing, databases~\cite{Motakis:1997:TAA:253260.253359}, and smart grid systems~\cite{erkin2013privacy}. 
%%%%%%%%%%%%%%%%%%%%%%%%

Unfortunately, the favorable properties of data aggregation come with major shortcomings. A plethora of data mining and machine learning tasks strive for data in finer granularity (disaggregated), thus data aggregation is undesirable. Along the same lines, algorithms designed for personalized analysis and accurate prediction significantly benefit from enhanced data resolution. Analysis results can differ substantially when using aggregated versus disaggregated data. Particularly, studies in the field of economics show that data aggregation results in information loss
and misleading conclusions at the individual level \cite{clark1976effects, garrett2003aggregated}. 
{Furthermore, in supply chain management, researchers have concluded that aggregating sales over time, products, or locations has a negative impact on demand forecasting \cite{jin2015masking}.}
On the other hand, disaggregation prior to analysis is very effective in environmental studies \cite{lenzen2011aggregation}, and leads to richer findings in learning analytics \cite{cole2010effects}. 
%%%%%%%%%%%%%%%%%%%%%%%%

The previous discussion reveals a clear {\em trade-off} between the need for data aggregation and the benefit of disaggregated data. This has motivated numerous works in developing algorithms for data disaggregation. In general, the task of data disaggregation is an inverse ill-posed problem. 
In order to handle the problem, classic techniques exploit side information or domain knowledge, in their attempt to {make the problem {overdetermined} and consequently} enhance the disaggregation accuracy. Some common prior models, imposed on the target higher resolution data, involve smoothness, periodicity \cite{liu2017h}, and non-negativity plus sparsity over a given dictionary \cite{almutairi2018homerun}. Such prior constraints are invoked when no other information is available about the data to be disaggregated.  

An interesting question arises when a dataset is aggregated over more that one dimension.
This is a popular research problem in the area of business and economics going back to the 70's \cite{rossi1982note, chow1971best}. 
In this case temporal {\em and} contemporaneous aggregated data are available \cite{pavia2010survey}.
For instance, given a country consisting of regions, we are interested in estimating the quarterly gross regional product (GRP) values, given the annual GRP per region (temporal aggregate) {\em and} the quarterly national series (contemporaneous aggregate) \cite{pavia2007estimating}. 
Another notable example appears in healthcare, where data are collected by national, regional, and local government agencies, health and scientific organizations, insurance companies and other entities, and are often aggregated in many dimensions (e.g., temporally, geographically, or by groups of hospitals), often to preserve privacy \cite{park2014ludia}.

Algorithms have been developed to integrate the multiple aggregates in the disaggregation task~\cite{rossi1982note, chow1971best, pavia2010survey, pavia2007estimating, di1990estimation}.  
The general disaggregation problem is ill-posed, which is clearly undesirable, even with multiple aggregates. Therefore, the majority of these works resort to adopting linear regression models with priors and additional information. However, it is unclear whether these formulations can identify the true disaggregated dataset under reasonable conditions. In this context, identifiability has not received the attention it deserves, likely because guaranteed recovery is considered mission impossible under realistic conditions. With multiview data aggregated in different ways, however, the problem can be well-posed, as we will show in this paper. 

\begin{figure}[t]
	\centering
	\includegraphics[width=3.5cm,height=2.75cm]{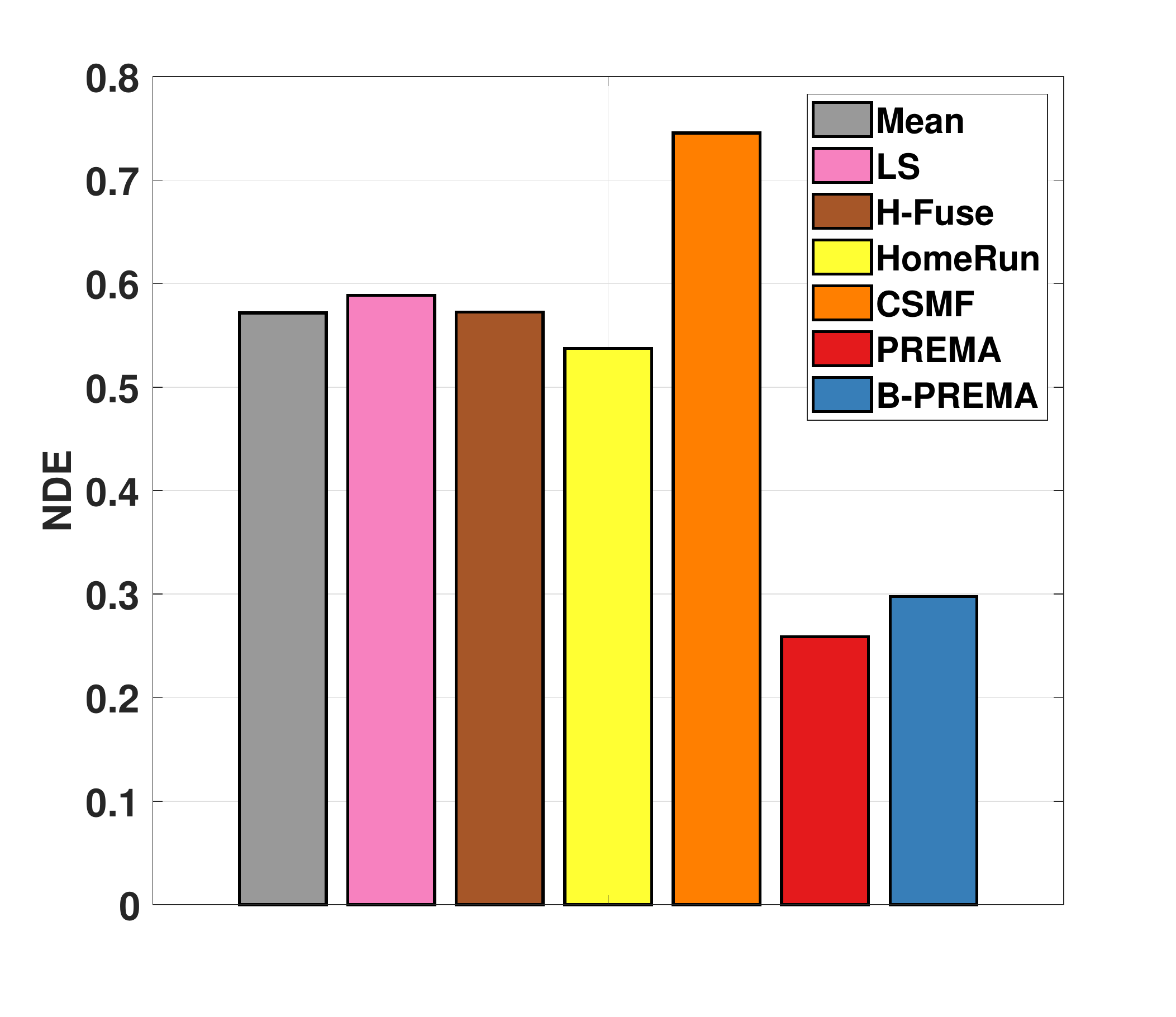}
	\setlength{\belowcaptionskip}{-7pt}
	\vspace{-8pt}
	\caption{\tensorglue is effective with real data.}
	\label{fig:intro_results}
\end{figure}

Our work is inspired by the following question: {\em Is the disaggregation task possible when the data are: 1) multidimensional, and 2) observed by different agencies via diverse aggregation mechanisms?} This is a well motivated problem due to the ubiquitous presence of data with multiple dimensions (three or more), also known as tensors, in a large number of applications. Note that aggregation often happens in more than one dimensions {\emph{of the same data}} as in the previously explained examples. The informal definition of the problem is given as follows: 
\begin{infproblem}[Multidimensional Disaggregation] 
	~	\vspace{-2mm}
	\begin{itemize}
		\itemsep0em
		\item {\bf Given:} two (or more) observations of a multidimensional dataset, each representing a different coarse view of the same data aggregated in one dimension (e.g., temporal and contemporaneous aggregates). 
		\item {\bf Recover:} the data in higher resolution (disaggregated) in all the dimensions.
	\end{itemize}
	\vspace{-2mm}
\end{infproblem}
%%%%%%%%%%%%%%%%%%%%%%%%%%%%%%%%%%%%%%%%%%%%%%%%%

We propose \tensorglue: a framework for fusing the multiple aggregates of multidimensional data. The proposed approach represents the target high resolution data as a {\em tensor}, and models that tensor  using the {\em canonical polyadic decomposition} (CPD) to reduce the number of unknowns, while capturing correlations and higher-order statistical dependencies across dimensions. \tensorglue employs a coupled CPD approach and estimates the low-rank factors of the target data, to perform the disaggregation task. 
This way, the originally ill-posed disaggregation problem is transformed to an overdetermined one, by leveraging the uniqueness properties of the CPD. \tensorglue is flexible in the sense that it can perform the disaggregation task on partially observed data, or data with missing entries. This is practically important as partially observed data appear often in real-world applications.

Our \tensorglue approach takes into account several well-known challenges that often emerge in real-life databases: the available measurements can have different scales (e.g., mixed monthly and yearly sums), gaps in the timeline (i.e., periods with no value reported), or time overlap (i.e., periods covered by more that one measurement). We also propose a variant of \tensorglue called \blind that handles the disaggregation task in cases where the aggregation pattern is unknown. The proposed framework not only provides a disaggregation algorithm, but it also gives insights that can be exploited in creating more  accurate data summaries for database applications. Interestingly, our work also provides insights on when aggregation {\em does not} preserve anonymity.  
%%%%%%%%%%%%%%%%%%%%%%%%%%%%%%%

We evaluated \tensorglue on real data from different domains, i.e., retail sales, crime counts, and weather observations. 
Experimental results show that the proposed algorithm reduces the disaggregation error of the best baseline by up to {67\%}.
Figure \ref{fig:intro_results} shows the Normalized Disaggregation Error (NDE) of \tensorglue and the baselines {with real data of the weekly sales counts of items in different stores of a retail company (CRA dataset, described in Section \ref{sec:data})}. We are given two observations: {1) monthly sales aggregates per store, and 2) weekly sales aggregated over groups of stores ($94$ stores are geographically divided into $18$ areas).}
\tensorglue outperforms all the competitors, even if the disaggregation pattern is unknown (\blind)---all the baselines use the aggregation information.   
The fact that the naive mean (Mean) gives a large error, indicates that {the data are not smooth and} the task is difficult. 
%%%%%%%%%%%%%%%%%%%%%%%%%%%%%%%%%%%

In summary, the contributions of our work are: 
\vspace{-2pt}
\begin{itemize}
	\vspace{-1pt}
	\item {\bf Formulation}: we formally define the multidimensional data disaggregation task from multiple views, aggregated across different dimensions, and provide an efficient algorithm.  
	\item {{{\bf Identifiability:}}} the considered model can provably transform the original ill-posed disaggregation problem to an identifiable one.
	\item {\bf Effectiveness:} \tensorglue recovers data with {large} improvement over the competing methods on real data.
	\item {\bf Unknown aggregation:} {the proposed model works even when the aggregation mechanism is unknown.}
	\item {\bf Flexibility :} {\tensorglue can disaggregate partially observed data.} 
	%\eit
	\vspace{-1pt}
\end{itemize}
%\vspace{-3pt}

\noindent{\bf Reproducibility:} The datasets we use are publicly available; our code is also available online\footnote{Code is available in https://github.com/FaisalAlmutairi/Prema}.

{Preliminary results of part of this work were presented in the {\em Proceedings of the Pacific-Asia Conference on Knowledge Discovery and Data Mining (PAKDD) 2020} \cite{Tendi}. 
In this journal version, the problem formulation is generalized to handle aggregated data with missing entries. Although accounting for missing entries makes the problem more complicated, our proposed models and careful algorithmic design yield an algorithmic framework that is efficient and comparable to \cite{Tendi} (which does not handle missing entries), both in terms of accuracy and computational complexity. We also provide identifiability proofs, detailed model and complexity analysis, and conduct extensive experiments.}

The rest of the paper is structured as follows. We explain the needed background and the related work in Section \ref{sec:back}, and introduce our proposed method in Section \ref{sec:proposed}. Then, we explain our experimental setup in Section \ref{sec:design} and show the experimental results in Section \ref{sec:results}. Finally, we summarize conclusions and take-home points in Section~\ref{sec:con}.
%%%%%%%%%%%%%%%%%%%%%%%%%%%%%%%%%%%%%%%%%%%%%%%%%%%%%%%%%%%%%%%%%%%%%%
\begin{table}[!]%[htbp]%[!htb]
	\caption{\small Symbols and Definitions}
	\vspace{-4pt}
	\centering
	\label{table:notation}
	\setlength\belowcaptionskip{-55pt}
	\resizebox{0.35\textwidth}{!}{
		\begin{tabular}{c| c }
			\hline	
			\hline
			Symbol &  Definition \\
			\hline	
			${\bf x}$, ${\bf X}$, $\tensorX$ & Vector, matrix, tensor\\
			${\bf X}_{n}$ & Mode-n matricization (unfolding)\\
			$\|.\|_F$ & Frobenius norm of a matrix/tensor\\
			${\bf X}^T$ & Transpose of matrix ${\bf X}$\\ 
			vec(.) & Vectorization operator of a matrix/tensor\\ 
			$[\![.]\!]$ & Kruskal operator, e.g., $\tensorX \approx [\![{\bf A}, {\bf B},{\bf C}]\!]$\\
			$\circ$ & Outer product\\
			$\otimes$ & Kronecker product\\
			$\odot$ & Khatri-Rao product (column-wise Kronecker)\\
			$\circledast$ & Hadamard (element-wise) product\\
			\hline
			\hline
	\end{tabular}}
\vspace{-2mm}
\end{table} 
\section{Background \& RELATED WORK }
\label{sec:back}
In this section, we review some tensor algebraic tools utilized by the proposed framework, define the disaggregation problem, 
and provide an overview of the related work.
Table \ref{table:notation} summarizes the main symbols and operators used throughout the paper.

\subsection{Tensor Algebra}
Tensors are multidimensional arrays indexed by three or more indices, $(i,j,k,...)$. A third-order tensor $\tensorX \in \R^{I \times J \times K}$ consists of three modes: rows $\tensorX(:,j,k)$, columns $\tensorX(i,:,k)$, and fibers $\tensorX(i,j,:)$.
Moreover, $\tensorX(i,:,:)$, $\tensorX(:,j,:)$, and $\tensorX(:,:,k)$ denote the $i^{th}$ horizontal, $j^{th}$ lateral, and $k^{th}$ frontal slabs of $\tensorX$, respectively.

\vspace{3pt}
\noindent {\bf Tensor decomposition (CPD/PARAFAC):} 
The outer product of two vectors $({\bf a} \circ {\bf b})$ results in a rank-one matrix. A rank-one third-order tensor $\tensorX \in \R^{I \times J \times K}$ is an outer product of three vectors: $\tensorX(i,j,k) = {\bf a}(i){\bf b}(j){\bf c}(k),~\forall i\in\{1,...,I\},~j\in\{1,...,J\},$ and $k\in\{1,...,K\}$, i.e., $\tensorX = ({\bf a} \circ {\bf b} \circ {\bf c})$, where ${\bf a}\in \R^I$, ${\bf b}\in \R^J$, and ${\bf c}\in \R^K$. The Canonical Polyadic Decomposition (CPD) (also known as PARAFAC) of a third-order tensor $\tensorX\in \R^{I \times J \times K}$ decomposes it into a sum of $R$ rank-one tensors \cite{hitchcock1927expression}, i.e., 
\begin{equation}\label{eq:CPD}
\tensorX = \sum_{r=1}^{R} {\bf a}_r \circ {\bf b}_r \circ {\bf c}_r
\end{equation}
where $R$ is the {\em tensor rank} and represents the minimum number of outer products needed, and ${\bf a}_r \in\mathbb{R}^{I}$, ${\bf b}_r \in\mathbb{R}^{J}$, and ${\bf c}_r \in\mathbb{R}^{K}$.
For brevity, we use $\tensorX = [\![{\bf A}, {\bf B},{\bf C}]\!]$ to denote the relationship in \eqref{eq:CPD}.$~{\bf A}\in\mathbb{R}^{I\times R}$, ${\bf B} \in\mathbb{R}^{J\times R}$, and ${\bf C} \in\mathbb{R}^{K\times R}$ are the factor matrices with columns ${\bf a}_r$, ${\bf b}_r$ and ${\bf c}_r$ respectively, i.e., ${\bf A} = [{\bf a}_1~~{\bf a}_2\dots{\bf a}_R]$, and likewise for ${\bf B}$ and ${\bf C}$.

\vspace{3pt}
\noindent\textbf{CPD uniqueness:}
An important property of the CPD is that ${\bf A},~{\bf B},~{\bf C}$ are essentially unique under mild conditions. CPD identifiability is established by the following theorem:
\begin{theorem}\label{thm:tensor_1}
	\cite{chiantini2012generic} 
	Let $\tensorX = [\![{\bf A},{\bf B},{\bf C}]\!]$ with ${\bf A} : I\times R$, ${\bf B} : J\times R$, and ${\bf C} : K\times R$. Assume $I\geq J\geq K$ without loss of generality. If $R \leq \frac{1}{16}JK$, then the decomposition of $\tensorX$ in terms of $\bf A, \bf B$, and $\bf C$ is essentially unique, almost surely -- i.e., for almost every (${\bf A}$, ${\bf B}$, ${\bf C}$) except for a set of Lebesgue measure zero.
\end{theorem}

Essential uniqueness means that ${\bf A},~{\bf B},~{\bf C}$ are unique up to common column permutation and scaling (scaling a column of one matrix that is compensated by counter-scaling the corresponding column of another matrix).

The CPD is also essentially unique, even if the tensor is incomplete (has missing entries). Several results have established CPD identifiability of tensors with missing entries, considering fiber sampled \cite{sorensen2019fiber}, regularly sampled \cite{kanatsoulis2019tensor} or randomly sampled tensors \cite{ashraphijuo2017fundamental}. The conditions for uniqueness are in general stricter compared to the case where the full tensor is available.

\vspace{3pt}
\noindent {\bf Tensor matricization (unfolding):}
There are three different ways to unfold (obtain a matrix view of) a third-order tensor $\tensorX$ of size $I \times J \times K$. First, the mode-3 unfolding is obtained by the vectorization and parallel stacking of the frontal slabs $\tensorX(:,:,k)$ as follows \cite{sidiropoulos2017tensor}
\begin{equation}\label{eq:mode3}
{\bf X}_{3} = [\text{vec}(\tensorX(:,:,1)),..., \text{vec}(\tensorX(:,:,K))] \hspace{10pt} \in \R^{IJ \times K}
\end{equation}
Equivalently, we can express ${\bf X}_{3}$ using the CPD factor matrices as ${\bf X}_{3} = ({\bf B} \odot {\bf A}){\bf C}^T$.
In the same vein, we may consider horizontal slabs to express the matricization over the first mode
\begin{equation}\label{eq:mode1}
\begin{aligned}
{\bf X}_{1} := & \hspace{2pt} [\text{vec}(\tensorX(1,:,:)),..., \text{vec}(\tensorX(I,:,:))]\\
= & \hspace{2pt} ({\bf C} \odot {\bf B}){\bf A}^T \in \R^{JK \times I}
\end{aligned}
\end{equation}
{or lateral slabs to obtain mode-2 unfolding}
\begin{equation}\label{eq:mode2}
\begin{aligned}
{\bf X}_{2} := & \hspace{2pt} [\text{vec}(\tensorX(:,2,:)),..., \text{vec}(\tensorX(:,J,:))]\\
= & \hspace{2pt} ({\bf C} \odot {\bf A}){\bf B}^T \in \R^{IK \times J}
\end{aligned}
\end{equation}

\begin{figure}[t]
	\centering
	\includegraphics[width = 0.31\textwidth,height=2.3cm]{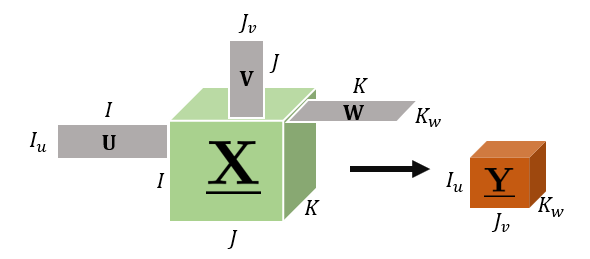}
	\vspace{-3pt}
	\setlength{\belowcaptionskip}{-7pt}
	\caption{\small Illustration of mode product with ($I_u < I$), ($J_v < J$), and ($K_w < K$).}
	\label{fig:mode_product}
\end{figure}

\vspace{3pt}
\noindent {\bf Mode product:}
It is the operation of multiplying a tensor by a matrix in one particular mode, e.g., mode-1 product of matrix ${\bf U}\in \R^{I_u \times I}$ and tensor $\tensorX\in\R^{I \times J \times K}$  corresponds to multiplying every column $\tensorX(i,:,k)$ of the tensor by ${\bf U}$~\cite{kolda2009tensor}. Similarly, mode-2 (mode-3) product corresponds to multiplying every row (fiber) of $\tensorX$ by a matrix.
Applying mode-1, mode-2, and mode-3 products to a third-order tensor $\tensorX\in\R^{I \times J \times K}$ jointly is represented using the following notation:
\begin{equation}\label{eq:mode_product}
\tensorY = \tensorX \times_1 {\bf U} \times_2 {\bf V} \times_3 {\bf W} \hspace{10pt} \in\R^{I_u \times J_v \times K_w}
\end{equation}
where ``$\times_n$" denotes the product over the $n^{th}$ mode, ${\bf U} \in \R^{I_u \times I}$, ${\bf V} \in \R^{J_v \times J}$, and ${\bf W} \in \R^{K_w \times K}$. 
Mode-1 multiplication results in reducing the tensor size in the first dimension {\em if} ($I_u < I$), similarly with the other modes; see Fig. \ref{fig:mode_product}. Moreover, 
if rows of ${\bf U}$ are binary vectors with more than one $1$, then each horizontal slab of $\tensorY$ is
{a sum of horizontal slabs of $\tensorX$ that correspond}
to the $1$'s in a particular row in ${\bf U}$. In the same vein, ${\bf V}$ and ${\bf W}$ could aggregate the lateral and frontal slabs, respectively. The mode product is also reflected in the CPD of the tensor, i.e., if ${\bf \tensorX}$ in the operation in \eqref{eq:mode_product} admits $\tensorX = [\![{\bf A}, {\bf B},{\bf C}]\!]$, then $\tensorY = [\![{\bf UA}, {\bf VB},{\bf WC}]\!]$.

\subsection{Disaggregation Problem}
\label{sec:dis}
The goal of the disaggregation task is to estimate a particular dataset in a higher resolution, given observations in lower resolution. In this subsection we present a high level linear algebraic view of disaggregation. This reveals the challenge of the task, which is the relationship between equations versus unknowns; detailed analysis follows in the next section. 

In the disaggregation problem, we are given a set of measurements ${\bf y} \in \R^{I_u}$ aggregated over the dataset ${\bf x} \in \R^{I}$, and our goal is to find $\bf x$. This can be cast as a linear inverse problem $\bf y= \bf U\bf x$, where ${\bf U} \in \R^{I_u \times I}$ is a `fat' {\em aggregation matrix} that relates the measurements to the unknown variables.
In this work, we consider the case where the target high-resolution data are multidimensional (tensor). 
Specifically, let $\tensorX\in\R^{I \times J \times K}$ be the target high-resolution third-order tensor. 
In the considered problem, we are given two sets of observations, each aggregated over one or more different dimension(s). This is common when data are reported by different agencies, resulting in multiple views of the same information.
The key insight is that the given aggregates can be modeled as mode product(s) of $\tensorX$ by an aggregation matrix in a particular mode(s). To see this, consider tensor $\tensorX\in\R^{4 \times 2 \times 2}$, a simple example of a set of observations aggregated over the first mode can be expressed as

%\vspace{-1mm}
{\small \begin{gather}
\underbrace{\begin{bmatrix}
{1}  & 1 & 0 & 0 \\ 
0  & {0} & 1 & 1 \\
\end{bmatrix}}_{{\bf U} \in \R^{2 \times 4}}
\times
\underbrace{\begin{bmatrix}\nonumber
x_{111} & x_{121} & x_{112} & x_{122} \\
x_{211} & x_{221} & x_{212} & x_{222}\\
x_{311} & x_{321} & x_{312} & x_{322}\\
x_{411} & x_{421} & x_{412} & x_{422}\\
\end{bmatrix}}_{{\bf X}_1^T\in \R^{4 \times (2 \times 2)}}\\
~~~~~~~~=
\underbrace{\begin{bmatrix}\label{pf}
y_{111} & y_{121} & y_{212} & y_{122}\\
y_{211} & y_{221} & y_{212} & y_{222}\\
\end{bmatrix}}_{{\bf Y}_1^T \in \R^{2 \times (2 \times 2)}}
\end{gather} \par}

\noindent where ${\bf X}_1$ and ${\bf Y}_1$ are mode-1 unfolding of ${\bf \tensorX}$ and ${\bf \tensorY}$, respectively. 
The same idea applies when the aggregation is over the second (third) mode using mode-2 (mode-3) product, respectively. 
In practical settings, the number of available aggregated measurements is much smaller than the number of variables (i.e., $I_u \ll I$), resulting in an under-determined, ill-posed problem. This is the major challenge of disaggregation, even when more than one set of aggregates are available.
An even more challenging case appears when one of the available {observation sets} is aggregated over more than one mode/dimension simultaneously (e.g., $\tensorY\in\R^{I_u \times J_v \times K}$, where $I_u < I$ and $J_v < J$ ). For instance, sales are reported for categories rather than individual items \emph{and} over groups of stores. This is a double aggregation over stores and items, and the proposed method can work under such a challenging scenario.
{Moreover, the aggregated observations might be partially observed (i.e., ${\bf Y}_1$ in \eqref{pf} has missing entries). This makes the problem more complicated, however our approach efficiently handles data with missing entries.} 

\subsection{Related Work}
\noindent {\bf Data disaggregation and fusion:}
The problem of data integration and fusion \cite{lenzerini2002data, dong2009data} from multiple sources has attracted the attention of several communities, due to the increasing access to all kinds of data, especially in database applications. A very challenging task in data integration, is that of
recovering a sequence of events (e.g., time series) from multiple aggregated reports \cite{faloutsos1997recovering, sax2013temporal, almutairi2018homerun,yang2020turbolift}. 
A common approach is to formulate the problem as linear least squares as in \eqref{pf}. In practice, however, the number of available aggregated samples is often significantly smaller than the length of the target series, resulting in an under-determined system of equations.  
To resolve this, previous algorithms have resorted to Tikhonov-type regularization of the ill-posed problem to impose some domain knowledge constraints, e.g., smoothness and periodicity \cite{liu2017h}. 

Fusing multiple observations aggregated in different dimensions for disaggregation purposes is a well studied task in the field of finance and economics \cite{rossi1982note, chow1971best, pavia2010survey, pavia2007estimating, di1990estimation}.
The considered approaches try to exploit linear relations between the target series in high resolution and the available aggregated measurements. However, this results in an under-determined linear system, even with multiple aggregates. Therefore, the majority of these works assume linear regression models with priors and additional information. Moreover, it is unclear whether the assumed models are identifiable, i.e., the model is not guaranteed to disaggregate the data. 

\noindent {\bf (Coupled) tensor  factorization:} Time series analysis, for various applications, is moving towards modern high-dimensional methods. For example, matrix and tensor factorization have been used in demand forecasting \cite{yu2016temporal}, mining and information extraction from complex time-stamped series \cite{matsubara2012fast}, and prediction of unknown locations in spatio-temporal data \cite{takeuchi2017autoregressive}.   

Data share common dimension(s) in a wide spectrum of applications. In such cases, coupled factorization techniques are commonly used to fuse the information for various objectives. For example, coupled factorization is often employed to integrate contextual information into the main data~\cite{almutairi2017context}. In recommender systems, for instance, we have a (user $\times$ item $\times$ time) tensor {\em and} a (user $\times$ features) matrix. In this case, the tensor and the features matrix are coupled in the user mode \cite{Papalexakis:2016:TDM:3004291.2915921}. Coupled tensor factorization has also been proposed for image processing \cite{kanatsoulis2018icip}, remote sensing \cite{kanatsoulis2018icassp}, and medical imaging problems \cite{kanatsoulis2019regular,kanatsoulis2019tensor}. Closest to our work is the approach in \cite{kanatsoulis2018hyperspectral}, which employs a coupled CPD to fuse a hyperspectral image with a multispectral image, to produce a high spatial and spectral resolution image. 
To our knowledge, this work {and its conference version \cite{Tendi} are the first that propose a tensor factorization approach} to tackle data disaggregation applications.

\begin{figure*}[t]
	\centering
	\includegraphics[width = 13cm,height=5.2cm]{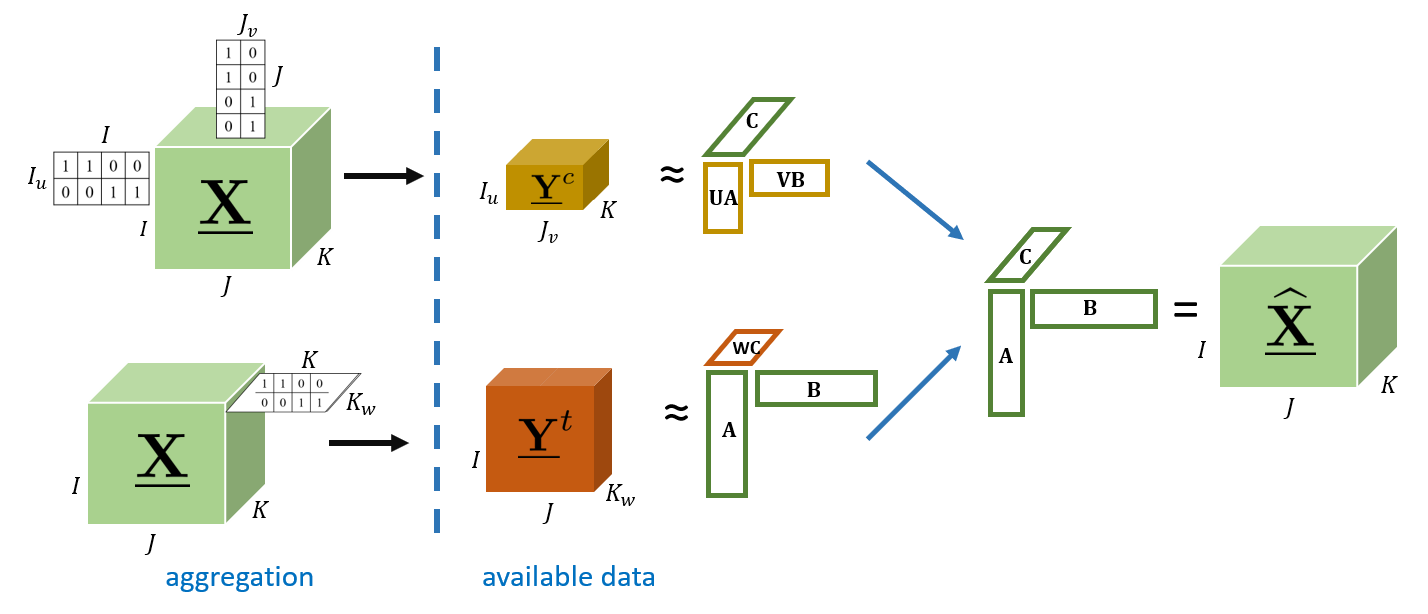}
	\setlength{\belowcaptionskip}{-12pt}
	\vspace{-2mm}
	\caption{Overview of \tensorglue.}
	\label{fig:TensorGlue}
\end{figure*}
\section{Proposed Framework: \tensorglue}
\label{sec:proposed}

Multidimensional data are indexed by multiple indices, e.g., $(i, j, k)$. Therefore, they can naturally be represented as a tensor $\tensorX \in \mathbb{R}^{{I} \times J \times K}$. The different modes represent the physical dimensions of the data (e.g., time stamps, locations, items, users). For the sake of simplicity of exposition, we focus on three-dimensional data in our formulations and algorithms. However, the proposed framework can handle more general cases with data of higher dimensions. 

In the remainder of this section, we give a detailed description and analysis of \tensorglue. Particularly, we state the problem and explain the proposed model in high level in Section \ref{sec:overview}, 
formulate \tensorglue in Section \ref{sec:formulation}, and present the main algorithm in Section \ref{sec:algorithm}. We discuss the complexity of \tensorglue  in Section \ref{sec:complexity}, and identifiability in section \ref{sec:ID}. Finally, we introduce \blind in Section \ref{sec:blindcase}, to tackle the disaggregation problem in the case where the aggregation matrices are unknown.

\subsection{Problem \& Model Overview}
\label{sec:overview}
Multidimensional aggregation is common when data are collected or released by different agencies, resulting in multiple views of the same dataset. 
We will explain the concept with the example of retail sales, which we use in the experiments. 
Estimating the retail sales in higher resolution enables accurate forecasting of future demand, and planing of economically efficient commerce. 
There are two sources of data used for this forecasting task:
1) Point of Sale (POS) data at the store-level, commonly aggregated in time (temporal aggregate $\tensorYt$);
and
2) historical {orders} made to the suppliers by the retailers' Distribution Centers (DC orders), aggregated over their multiple stores (contemporaneous aggregate $\tensorYc$).
In particular, DC  order data are immediately available to the suppliers, whereas the POS data are owned by the retailers. Both DC order and POS data are used to forecast demand, and especially POS data are vital in predicting future orders \cite{williams2010creating}. For that reason, many retailers share POS with their suppliers to assist in forecasting orders and avoid shortage or excess in inventory~\cite{jin2015forecasting}.
In a more restricted scenario, the second source collects information about each category of items rather than each item individually. 
{Oftentimes, data are partially observed, i.e., $\tensorYt$ and $\tensorYc$ have missing entries. In this example, not all items are offered in all stores during all the considered time stamps.  }
The question that arises is whether we can fuse these sources to reconstruct high-resolution data in stores, items, and time dimensions.

Formally, we are interested in the following:
\begin{problem}[{\bf Multidimensional Disaggregation}] 
	\begin{itemize}
		\itemsep0em
		\vspace{-2pt}
		\item {\bf Given} two aggregated views of three-dimensional data $\tensorX\in\R^{I \times J \times K}$: $\tensorYt\in\R^{I \times J \times K_w}$, and $\tensorYc\in\R^{I_u \times J \times K}$ (or $\tensorYc\in\R^{I_u \times J_v \times K}$), with $I_u < I$, $J_v < J$, and $K_w < K$, and possibly missing entries.
		\item \bf Recover the original disaggregated multidimensional data $\tensorX\in\R^{I \times J \times K}$.
	\end{itemize}
\end{problem}

Note that each aggregated view is the result of the mode product of the target data with an aggregation matrix. In particular $\tensorYt = \tensorX \times_3 {\bf W}$, where ${\bf W} \in \R^{K_w \times K}$ is an aggregation matrix with $K_w<K$,  
and $\tensorYc = \tensorX \times_1 {\bf U}$, where ${\bf U} \in \R^{I_u \times I}$ is an aggregation matrix with $I_u<I$. In the case where one view is jointly aggregated in 2 dimensions, e.g., sales are aggregated over groups of stores and groups of items,
$\tensorYc = \tensorX \times_1 {\bf U} \times_2 {\bf V}$,  where ${\bf V} \in \R^{J_v \times J}$ is an aggregation matrix with $J_v < J$.

\tensorglue aims to fuse the different available aggregates in order to estimate the multidimensional data in the desired higher resolution. 
At a higher level, 
the main idea behind the proposed method is that the target multidimensional data, 
$\tensorX\in\R^{I\times J\times K}$, admits a CPD model. Therefore, it can be well approximated using its CPD factors ${\bf A}, {\bf B}, {\bf C}$ (i.e., $\tensorX = [\![{\bf A}, {\bf B},{\bf C}]\!]$).
Exploiting the low-rank modeling helps in reducing the number of unknown variables, especially if the data are highly correlated. 
Then, the CPD factors of the two aggregated observations are
\begin{align}
\tensorYt = [\![{\bf A}, {\bf B},{\bf WC}]\!] \label{yt}\\
\tensorYc = [\![{\bf UA}, {\bf VB},{\bf C}]\!] \label{yc}
\end{align}
\tensorglue learns the factor matrices ${\bf A}$, ${\bf B}$, and ${\bf C}$ by applying a coupled CPD model on the available aggregates with respect to the available observations. 
{Note that up to this point, we have not explained how missing entries in $\tensorYt$ and $\tensorYc$ are treated, which will be discussed in the next section.}
Figure \ref{fig:TensorGlue} illustrates the high level picture of our model.

\subsection{\tensorglue: Formulation}
\label{sec:formulation}
If we have the original (disaggregated) data in the tensor {$\tensorX$ with missing entries, a common way to estimate its CPD factors is by adopting a least squares criterion to minimize the difference between the original tensor $\tensorX$ and its CPD $[\![{\bf A}, {\bf B},{\bf C}]\!]$ with respect to the available (observed) entries. This can be done by adding a weight tensor that masks the available entries, i.e.,}

\begin{equation}\label{eq:CPD_mode1}
\begin{aligned}
{\minimize_{{\bf A},{\bf B},{\bf C}}} \quad & \|{\bf \tensorW} \circledast ({\tensorX} -  [\![{\bf A}, {\bf B},{\bf C}]\!])\|_F^2
\end{aligned}
\end{equation}

\noindent where ${\bf \tensorW}$ is defined as 
\begin{equation}\label{eq:W}
{\bf \tensorW}(i,j,k) = 
\begin{cases} 1, &\mbox{if } \tensorX(i,j,k)~ \mbox{is available}\\
0, & \mbox{otherwise}
\end{cases}
\end{equation}
Fortunately, many real life data exhibit low-rankness due to the correlation between the elements within each dimension (e.g., stores, items, time stamps), i.e., $R$ in \eqref{eq:CPD} is small relative to the size of the tensor.   

In the considered disaggregation task, we only have aggregated views of the multidimensional data (i.e., compressed version of the target tensor $\tensorX$). {These aggregated views can have missing elements for various application-specific reasons such as privacy, lack of data collection, or absence of events.} 
We use the fact that the aggregated tensors share the same factors (up to aggregation) as shown in equations \eqref{yt} and \eqref{yc} to jointly decompose $\tensorYt$ and $\tensorYc$ by means of coupled tensor factorization. To this end, we obtain the following formulation:      
\begin{equation}\label{eq:CPD_Factors_agg13}
\begin{aligned}
\min_{{\bf A},{\bf B},{\bf C}} \quad & \mathcal{F}({\bf A}, {\bf B}, {\bf C}) := \|\tensorWt \circledast (\tensorYt -  ([\![{\bf A}, {\bf B},{\bf WC}]\!])\|_F^2\\
&\quad \quad
+ \|\tensorWc \circledast ({\tensorYc} -  ([\![{\bf UA}, {\bf VB},{\bf C}]\!])\|_F^2
\end{aligned}
\end{equation}
{where $\tensorWt \in \{0,1\}^{I \times J \times K_w}$ and $\tensorWc \in \{0,1\}^{I_u \times J_v \times K}$ are weight tensors with ones at the indices of the available entries in $\tensorYt$ and $\tensorYc$, respectively, and zeros elsewhere. As a result, the CPD factors $\bf{A}$, $\bf{B}$, and $\bf{C}$ are learned with respect to the available data.}
One could add a regularization parameter $\lambda$ to control the weight between the two terms, however, we observed that it does not significantly affect the disaggregation performance.
Note that if we have additional aggregated observations, we can incorporate them using the same concept. {Enforcing non-negativity constraints on the factors seems natural \emph{if} we are dealing with count data, however, we empirically observed that it does not improve the disaggregation accuracy.}

\subsection{\tensorglue: Algorithm}
\label{sec:algorithm}
The optimization in \eqref{eq:CPD_Factors_agg13} is non-convex, and NP-hard in general. To tackle it, we derive a \emph{Block Coordinate Descent} (BCD) algorithm that updates the three variables in an alternating fashion. 
Starting from initial factors ${\bf A}^{(0)}$, ${\bf B}^{(0)}$, and ${\bf C}^{(0)}$, at every iteration $k \in \mathbb{N}$, we cyclically update each factor while fixing the other two.
Each update is a step in the direction of the negative gradient of $\mathcal{F}$ with respect to the corresponding factor. To simplify the expressions, let us define ${\bf \widetilde{A}} = {\bf UA}$, ${\bf \widetilde{B}} = {\bf VB}$, and ${\bf \widetilde{C}} = {\bf WC}$. 
The partial derivative of the above objective function $\mathcal{F}$ w.r.t. ${\bf A}$ is as follows---the derivations are deferred to Appendix A. %\ref{sec:AppA}.
\begin{equation}\label{eq:updatA}
\begin{aligned}
\frac{\partial{\mathcal{F}}}{\partial{\bf A}} = \nabla_{\bf A}\mathcal{F} & = 2\big(\underbrace{{\bf \Omega}_1^{t} \circledast (({\bf \widetilde{C}} \odot {\bf B}){\bf A}^T - {\bf Y}_1^{t})}_{{\bf E}_t}\big)^T\big({\bf \widetilde{C}} \odot {\bf B}\big)\\
& + 2{\bf U}^T\big(\underbrace{{\bf \Omega}_1^{c} \circledast (({\bf C} \odot {\bf \widetilde{B}}){\bf \widetilde{A}}^T - {\bf Y}_1^{c})}_{{\bf E}_c}\big)^T\big({\bf C} \odot {\bf \widetilde{B}}\big).
\end{aligned}
\end{equation}
where {${\bf Y}_1^t$, ${\bf Y}_1^c$, ${\bf \Omega}_1^t$, and ${\bf \Omega}_1^c$} are mode-1 unfolding of the corresponding tensors. 
Similarly, we derive the derivatives of $\mathcal{F}$ w.r.t. ${\bf B}$ and ${\bf C}$ using mode-2 and mode-3 unfoldings of the tensors, respectively,
and get the following equations:
\begin{equation}\label{eq:updateB}
\begin{aligned}
\nabla_{\bf B}\mathcal{F} & = 2\big({\bf \Omega}_2^{t} \circledast (({\bf \widetilde{C}} \odot {\bf A}){\bf B}^T - {\bf Y}_2^{t})\big)^T\big({\bf \widetilde{C}} \odot {\bf A}\big)\\
& + 2{\bf V}^T\big({\bf \Omega}_2^{c} \circledast (({\bf C} \odot {\bf \widetilde{A}}){\bf \widetilde{B}}^T - {\bf Y}_2^{c})\big)^T\big({\bf C} \odot {\bf \widetilde{A}}\big),
\end{aligned}
\end{equation}

\begin{equation}\label{eq:updateC}
\begin{aligned}
\nabla_{\bf C}\mathcal{F} & = 2{\bf W}^T\big({\bf \Omega}_3^{t} \circledast ({\bf }({\bf B} \odot {\bf A}){\bf \widetilde{C}}^T - {\bf Y}_3^{t})\big)\big({\bf {B}} \odot {\bf {A}}\big)\\
& + 2\big({\bf \Omega}_3^{t} \circledast (({\bf \widetilde{B}} \odot {\bf \widetilde{A}}){\bf C}^T - {\bf Y}_3^c)\big)^T\big({\bf C} \odot {\bf \widetilde{A}}\big)
\end{aligned}
\end{equation}
With the above gradient expressions at hand, we have established the update direction for each block (factor), which is the negative gradient of $\mathcal{F}$ with respect to each factor:
\begin{align}
{\bf A} = {\bf A} - \alpha\nabla_{\bf A}\mathcal{F},\label{eq:A}\\
{\bf B} = {\bf B} - \beta\nabla_{\bf B}\mathcal{F},\label{eq:B}\\
{\bf C} = {\bf C} - \gamma\nabla_{\bf C}\mathcal{F}\label{eq:C}.
\end{align}
{We now seek to select the step-size terms $\alpha$, $\beta$, and $\gamma$. We use the \emph{exact line search} approach for this task. At every iteration $k\in\mathbb{N}$, $\alpha$ is chosen to minimize $\mathcal{F}$ along the line $\{{\bf A} - \alpha\nabla_{\bf A}\mathcal{F}|\alpha\geq0\}$}
\begin{equation}\label{eq:alpha}
\begin{aligned}
\argmin_{\alpha\geq 0} \quad & \mathcal{F}\big({\bf A} - \alpha\nabla_{\bf A}\mathcal{F}\big)
\end{aligned}
\end{equation}
Luckily, in our case, the above optimization can be solved optimally without extra heavy computations. The optimal solution to \eqref{eq:alpha} is as follows (refer to Appendix B 
%\ref{sec:AppB} 
for derivations). 
\begin{equation}\label{eq:alphasol}
\alpha  = max\big(0,\frac{ {\bf e}_t^T{\bf g}_t + {\bf e}_c^T{\bf g}_c}{{\bf g}_t^T{\bf g}_t + {\bf g}_c^T{\bf g}_c}\big),
\end{equation}
where ${\bf e}_t = \text{vec}({\bf E}_t)$, ${\bf e}_c = \text{vec}({\bf E}_c)$, with ${\bf E}_t$ and ${\bf E}_c$ are as defined in \eqref{eq:updatA}, and 
\begin{align}
{\bf g}_t & = \text{vec}({\bf \Omega}_1^{t} \circledast (({\bf \widetilde{C} \odot {\bf B}})\nabla_{\bf A}\mathcal{F}^T)),\label{gt}\\
{\bf g}_c & = \text{vec}({\bf \Omega}_1^{c} \circledast (({\bf {C}} \odot {\bf \widetilde{B}})({\bf U}\nabla_{\bf A}\mathcal{F})^T))\label{gc}.
\end{align}
Note that ${\bf E}_t$ and ${\bf E}_c$ are already computed in \eqref{eq:updatA}. We have also computed $({\bf \widetilde{C}}  \odot {\bf B})$ and $({\bf {C}}  \odot {\bf \widetilde{B}})$ in \eqref{eq:updatA}, which are needed to obtain ${\bf g}_t$ amd ${\bf g}_c$, respectively. 
Thus, the exact line search step only requires:
\begin{itemize}
    \item Multiplying the transpose of the gradient $\nabla_{\bf A}\mathcal{F} \in \R^{I \times R}$ by a $K_wJ \times R$ matrix in \eqref{gt} (and ${\bf U}\nabla_{\bf A}\mathcal{F} \in \R^{I_u \times R}$ by a $KJ_v \times R$ matrix in \eqref{gc}).
    \item Computing the inner products in \eqref{eq:alphasol}.
\end{itemize}

{In a similar fashion, $\beta$ and $\gamma$ are obtained by solving the following optimization functions, respectively:}
\begin{equation}\label{eq:beta}
\begin{aligned}
\beta =  \argmin_{\beta\geq 0} \quad & \mathcal{F}\big({\bf B} - \beta\nabla_{\bf B}\mathcal{F}\big)
\end{aligned}
\end{equation}
\begin{equation}\label{eq:gamma}
\begin{aligned}
\gamma =  \argmin_{\gamma\geq 0} \quad & \mathcal{F}\big({\bf C} - \gamma\nabla_{\bf C}\mathcal{F}\big)
\end{aligned}
\end{equation}
The solutions to the above are similar to the case of $\alpha$, but with mode-2 and mode-3 tensor unfoldings. We provide {an illustrative} example of deriving the solution to \eqref{eq:alpha}, \eqref{eq:beta}-\eqref{eq:gamma} in Appendix B.
%\ref{sec:AppB}. 
The overall steps of \tensorglue are summarized in Algorithm~\ref{alg:algo1}.

%\vspace{-2mm}
\begin{algorithm}[h!]%[htbp]%[!htb]
	\raggedright
	\caption{: \tensorglue \eqref{eq:CPD_Factors_agg13}}
	\label{alg:algo1}
	\textbf{input:} $\tensorYt$, $\tensorYc$, ${\bf U}$, ${\bf V}$, ${\bf W}$, $R$\\
	\textbf{Initialize:} ${\bf A}$, ${\bf B}$, ${\bf C}$ (refer to Appendix~~C)\\
	%\ref{sec:AppC})\\
	%\vspace{3pt}
	\textbf{Repeat}
	\vspace{-3pt}
	\begin{itemize}
		\itemsep0em
		\item Update ${\bf A}$ using \eqref{eq:A}, \eqref{eq:updatA}, and \eqref{eq:alphasol}
		\item  Update ${\bf B}$ using \eqref{eq:B}, \eqref{eq:updateB}, and \eqref{eq:beta}
		\item  Update ${\bf C}$ using \eqref{eq:C}, \eqref{eq:updateC}, and \eqref{eq:gamma}
	\end{itemize}
	\vspace{-3pt}
	\textbf{Until} criterion is met (max. $\#$iterations)\\
	\textbf{output:} ${\bf A}$, ${\bf B}$, ${\bf C}$\\
\end{algorithm}
%\vspace{-1.5mm}

We observed empirically that a careful initialization for the factor matrices in Algorithm~\ref{alg:algo1} results in a better disaggregation accuracy, and substantially reduces the operational time (i.e., reduces the required number of iterations). 
Thus, we design a careful initialization method based on CPD.
First, we set the missing entries to zero, then perform CPD on one tensor to get initial estimates of two factors. Then, we solve a system of linear equations using the other tensor to obtain an initial estimate of the third factor. For instance, from CPD($\tensorYt$) we get ${\bf A}$, ${\bf B}$, and ${\bf \widetilde{C}}$. Then, we obtain ${\bf C}$ by solving the linear system ${\bf Y}_3^c = \big(({\bf VB}) \odot ({\bf UA})\big){\bf C}^T$. This way, we establish an initial guess for ${\bf A}$, ${\bf B}$, and ${\bf {C}}$. We provide detailed initialization steps in Appendix~C.
%\ref{sec:AppC}.

\subsection{\tensorglue: {Complexity Analysis}}
\label{sec:complexity}
The complexity of \tensorglue is determined by the matrix multiplication operations required to obtain the gradients and the step size terms. The products in the gradient expressions have the dominant computational cost. Therefore, we break down the computational complexity below using the gradient w.r.t. ${\bf A}$ in \eqref{eq:updatA};
the complexity of computing the gradients w.r.t ${\bf B}$ and ${\bf C}$ are similar. Recall \eqref{eq:updatA}:
\begin{equation}\label{eq:updatA2}
\begin{aligned}
\nabla_{\bf A}\mathcal{F} & = 2\big(\underbrace{{\bf \Omega}_1^{t} \circledast (({\bf \widetilde{C}} \odot {\bf B}){\bf A}^T - {\bf Y}_1^{t})}_{{\bf E}_t \in \R^{JK_w \times I}}\big)^T\big({\bf \widetilde{C}} \odot {\bf B}\big)\\
& + 2{\bf U}^T\big(\underbrace{{\bf \Omega}_1^{c} \circledast (({\bf C} \odot {\bf \widetilde{B}}){\bf \widetilde{A}}^T - {\bf Y}_1^{c})}_{{\bf E}_c \in \R^{J_vK \times I_u}}\big)^T\big({\bf C} \odot {\bf \widetilde{B}}\big)\nonumber
\end{aligned}
\end{equation}

\begin{enumerate}
    \item  Computing the two Khatri-Rao products costs $\mathcal{O}(K_wJR + KJ_vR)$, where $R$ is the rank.
    
    \item The cost of multiplying $({\bf \widetilde{C}} \odot {\bf B})$ with ${\bf A}^T$, and $({\bf C} \odot {\bf \widetilde{B}})$ with ${\bf \widetilde{A}}^T$ is $\mathcal{O}(IJK_wR + I_uJ_vKR)$. 

    \item The element-wise products ($\circledast$) cost $\mathcal{O}(nnz(\tensorWt) + nnz({\tensorWc}))$.
    
    \item Multiplying ${\bf E}_t^T$ and ${\bf E}_c^T$ with the Khatri-Rao products costs
    $\mathcal{O}(R(nnz(\tensorWt) + nnz({\tensorWc})))$. 
    
    \item {In the worst case where ${\bf U}$ and ${\bf \Omega}_1^c$ have no zeros, the cost of multiplying ${\bf U}^T$ with ${\bf E}_c^T({\bf C} \odot {\bf \widetilde{B}})$ is $\mathcal{O}(II_uR)$.} 
\end{enumerate}

The dominant cost terms are in the $2^{nd}$ point above. Thus, the overall complexity is $\mathcal{O}(IJK_wR + I_uJ_vKR)$. Since $R$ is usually very small relative to the size of the tensors in real data, the complexity is linear with the size of $\tensorYt$ and $\tensorYc$.

\subsection{{\tensorglue: Identifiability Analysis}}
\label{sec:ID}
After introducing the model and the algorithm, we establish the identifiability of the \tensorglue model. As mentioned earlier, the multidimensional disaggregation task is an inverse ill-posed problem. Considering a low rank CPD model on the data, results in a tensor disaggregation problem with a unique solution. In other words, the optimal solution of \eqref{eq:CPD_Factors_agg13} is guaranteed to be unique, under mild conditions, and identify the original fine-resolution tensor almost surely. For the sake of simplicity we first assume that $\tensorYt$ does not have any missing values.

%\vspace{-1pt}
\begin{proposition}\label{thm:known}
	Let $\tensorX\in\R^{I \times J \times K}$ be the target tensor to disaggregate with CPD $\tensorX = [\![{\bf A}, {\bf B},{\bf C}]\!]$ of rank $R$. Also let $\tensorYt\in\R^{I \times J \times K_w} = \tensorX \times_3 {\bf W}$ 
	and $\tensorYc \in\R^{I_u \times J_v \times K} = \tensorWc \circledast(\tensorX \times_1 {\bf U}\times_2 {\bf V})$ be the two aggregated sets of observations. Assume that ${\bf A}$, ${\bf B}$ and ${\bf C}$ are drawn from some absolutely continuous joint distribution with respect to the Lebesque measure in $\mathbb{R}^{(I\times J\times K)R}$, and that $({\bf A}^\star,{\bf B}^\star,{\bf C}^\star)$ is an optimal solution to problem~\eqref{eq:CPD_Factors_agg13}. Also assume that the number of observed entries at each frontal slab of $\tensorYc$ is greater than or equal to $R$. Then, $\widehat{\tensorX} = [\![{\bf A}, {\bf B},{\bf C}]\!]$ disaggregates $\tensorYt,~\tensorYc$ to $\tensorX$ almost surely if $R \leq \frac{1}{16}\min\{IJ,IK_w,JK_w,16I_uJ_v\}$.
\end{proposition}

%\vspace{-2pt}
The proof is intuitive and parallels recent 
results obtained in the hyperspectral imaging literature \cite{kanatsoulis2018hyperspectral}.
\textbf{Proof sketch:} We use Theorem \ref{thm:known} to claim identifiability of $\tensorYt$. Then factors $\bf A,~\bf B$ can be identified up to common permutation and scaling. The solution for $\bf C$ is obtained via solving an overdetermined linear system of equations using $\tensorYc$. This way permutation and scaling is preserved and the target tensor is recovered as $\tensorX = [\![{\bf A}, {\bf B},{\bf C}]\!]$. In the case where $\tensorYt$ has missing entries, identifiability depends on the pattern of missings. Specifically, the results in \cite{sorensen2019fiber}, \cite{kanatsoulis2019tensor}, \cite{ashraphijuo2017fundamental} can be employed, when the available measurements are fiber, regularly or randomly sampled respectively. The conditions are more restrictive compared to the case of fully observed tensor, but guarantee identifiability of $\bf A,~\bf B$ up to common permutation and scaling. The solution for $\bf C$ is the same as in the previous case. The detailed proof is presented in Appendix D.
%\ref{sec:AppD}.
%%%%%%%%%%%%%%%%%%%%%%%%%%%%%%%%%%%%%%%%%

\subsection{\blind: \tensorglue with Unknown Aggregation}\label{sec:blindcase}

In most practical applications, the aggregation details are known. However, there exist cases with limited knowledge on how the data are aggregated, i.e., we do not know (or have partial knowledge of) ${\bf U}$, ${\bf V}$, and ${\bf W}$. {We consider the case where each available view is aggregated in one dimension, and propose the following formulation to get the factors of the disaggregated tensor (${\bf A}$, ${\bf B}$, and ${\bf C}$):}

\begin{equation}\label{eq:blind}
\begin{aligned}
\min_{\begin{subarray}{c}{\bf A},{\bf B},{\bf C}, {\bf \widetilde{A}}, {\bf \widetilde{C}}\end{subarray}}  & ~\mathcal{L}({\bf A},{\bf B},{\bf C}, {\bf \widetilde{A}}, {\bf \widetilde{C}}) := \|\tensorWt \circledast (\tensorYt -  [\![{\bf A}, {\bf B},{\bf \widetilde{C}}]\!])\|_F^2\\
& \quad \quad + \|\tensorWc \circledast (\tensorYc -  [\![{\bf \widetilde{A}}, {\bf {B}},{\bf C}]\!])\|_F^2 
+ \mu \mathcal{R}({\bf C}, {\bf \widetilde{C}}) 
\end{aligned}
\end{equation}
Where $\widetilde{\bf A} = {\bf UA}$, and $\widetilde{\bf C} = {\bf WC}$ are treated as separate variables since we do not know ${\bf U}$ and ${\bf W}$, {and $\mathcal{R}$ is a regularization function.} 
This problem is more challenging than \eqref{eq:CPD_Factors_agg13} as the number of variables has been increased, with the same number of equations.
Another challenge is that there is a scaling ambiguity between the factors of the two tensors {\emph{if} we omit the regularization term in \eqref{eq:blind}.}
{Scaling and counter-scaling the factors of the tensor $\tensorYt$ (or $\tensorYc$) does not change its estimated value, or the value of the cost function in \eqref{eq:blind}.   
For example, scaling ${\bf A}$ by a $\lambda$, and $\widetilde{\bf C}$ by $1/\lambda$ does not change the value of $\widehat{{\bf Y}}^t_1 = (\widetilde{\bf C} \odot {\bf B}){\bf A}^T$, and as a result, it gives the same cost value.
However, this scaling changes the estimated value of the disaggregated tensor $\widehat{\bf X}_1  = ({\bf C} \odot {\bf B}){\bf A}^T$. This is because tensor $\tensorX$ shares factors with both $\tensorYt$ and $\tensorYc$.}
To overcome this, we observe that the temporal aggregation ${\bf W}$ in most aggregated data is non-overlapping and includes all the time ticks\footnote{{\em Known} overlap, e.g., $50\%$, can be treated similarly -- as in this case every atom is counted twice.}. 
This means that the respective column sums of ${\bf C}$ and $\widetilde{\bf C}$ should be equal. We exploit this observation by choosing the following regularization term for \eqref{eq:blind}  
$$\mathcal{R}({\bf C}, {\bf \widetilde{C}}) = \|{\bf 1}^T{\bf C} - {\bf 1}^T{\bf \widetilde{C}}\|_2^2,$$
{which reconciles for the scaling ambiguity.} 

In order to tackle the problem above, we derive a BCD algorithm, in the same fashion as Algorithm~\ref{alg:algo1}. 
The steps are summarized in Algorithm \ref{alg:algo2}. We alternate between updating the five variables. In each update, we take a step in the direction of the negative gradient w.r.t. the corresponding variable. The derivations of the gradients are shown in Appendix A. %\ref{sec:AppA}. 
The step size parameters $\alpha, \rho, \beta, \gamma$, and $\sigma$ are chosen using the exact line search explained in section \ref{sec:algorithm} and Appendix B.
%\ref{sec:AppB}.     

To initialize the factors in Algorithm \ref{alg:algo2}, we set the missing entries to zero, then we use {\fontfamily{qcr}\selectfont Tensorlab} and compute ({\fontfamily{qcr}\selectfont CPD}($\tensorYc$)) to get ${\bf \widetilde{A}}$, ${\bf B}$, and ${\bf C}$. 
To get an initial estimate of ${\bf \widetilde{C}}$, 
{we exploit the fact that the temporal aggregates are the summation over consecutive time stamps in most real data.}
Therefore, we sum every consecutive $w = \frac{K}{K_W}$ rows in ${\bf C}$. This way we approximate the temporal aggregation process in a very intuitive way, the true aggregation matrix being unknown\footnote{In the experiments, we make sure that the true temporal aggregation and the estimated one do not align.}.

%OK ME

\section{Experimental Design}
\label{sec:design}
In this section, we provide a detailed description of the setup we use in our experiments. First, we describe the data used in the experiments. Then, we explain the aggregation applied on these data to generate aggregated views. Last, we present the evaluation metrics and baselines used for comparison.

\subsection{Datasets}
\label{sec:data}
We evaluate \tensorglue using {the following public datasets, which are readily available online:} 

\vspace{3pt}
\noindent {\bf DFF\footnote{{https://www.chicagobooth.edu/research/kilts/datasets/dominicks}}:} Retail sales data, called Dominick's Finer Foods (DFF),
collected by 
the James M. Kilts Center, University of Chicago Booth School of Business. 
DFF used to be a grocery store chain based in the Chicago area until all of its stores were closed.
Sales, in this dataset, are divided into category-specific files. In particular, each file contains the weekly sales (i.e., number of sold units) of items belonging to a specific category (e.g., cheese, cookies, soft drinks, etc) in about $100$ stores. DFF data contain the geographical locations of the different stores, which we use to aggregate stores into groups. 
We create ground truth three-dimensional tensors, using $10$ different category-specific datasets. This way, a (stores $\times$ items $\times$ weeks) tensor is formed \emph{for each category}. 
These 10 department-specific datasets are listed as the first group in Table \ref{table:data}---we use the three bold letters acronym for these categories in the results.
We pick the $50$ most popular items from each category. 
Note that this results in an `incomplete' tensor, owing to the fact that not all items were offered in all stores, or they were offered only for part of the time in some stores. These tensors have varying statistics (see Table \ref{table:data}), which allows thorough testing and analysis. We also form an additional (stores $\times$ items $\times$ weeks) tensor that contains items from all the $10$ different categories combined, $50$ items from each (namely {\bf Mixed DFF} in Table~\ref{table:data}).

\begin{algorithm}[t]%[htbp]%[!htb]
	\raggedright
	\caption{: \blind}
	\label{alg:algo2}
	\textbf{input:} $\tensorYt$, $\tensorYc$, $R$, $\mu$\\
	\textbf{Initialize:} $\widetilde{{\bf A}}$, ${\bf B}$, ${\bf C}$, $\leftarrow$ {\fontfamily{qcr}\selectfont CPD}($\tensorYc$)\\
	${\bf \widetilde{C}}(k_w,:) \leftarrow \sum_{k = w(k_w - 1) + 1}^{w \times k_w} {\bf C}(k,:)$ \\
	${\bf A} \leftarrow \text{solve}~ {\bf Y}_3^{t} = {\bf A}({\bf \widetilde{C}} \odot {\bf B})^T$\\
	%\vspace{3pt}
	\textbf{Repeat}
	\vspace{-3pt}
	\begin{itemize}
		\itemsep0em
	        \item $\alpha \leftarrow \argmin_{\alpha\geq 0} ~\mathcal{L}({\bf A}- \alpha \nabla_{\bf A}\mathcal{L})$; ~~${\bf A} = {\bf A} - \alpha\nabla_{\bf A}\mathcal{L}$
			\item $\rho \leftarrow \argmin_{\rho\geq 0} ~\mathcal{L}({\bf \widetilde{A}}- \rho \nabla_{\bf \widetilde{A}}\mathcal{L})$; ~~ ${\bf \widetilde{A}} = {\bf \widetilde{A}} - \rho\nabla_{\bf \widetilde{A}}\mathcal{L}$
		    \item $\beta \leftarrow \argmin_{\beta\geq 0} ~\mathcal{L}({\bf B}- \beta \nabla_{\bf B}\mathcal{L})$; ~~${\bf B} = {\bf B} - \beta\nabla_{\bf B}\mathcal{L}$
             \item $\gamma \leftarrow \argmin_{\gamma\geq 0} ~\mathcal{L}({\bf C}- \gamma \nabla_{\bf C}\mathcal{L})$; ~~${\bf C} = {\bf C} - \gamma\nabla_{\bf C}\mathcal{L}$
            \item $\sigma \leftarrow \argmin_{\sigma\geq 0} ~\mathcal{L}({\bf \widetilde{C}}- \sigma \nabla_{\bf \widetilde{C}}\mathcal{L})$; ~~${\bf \widetilde{C}} = {\bf \widetilde{C}} - \sigma\nabla_{\bf \widetilde{C}}\mathcal{L}$
	\end{itemize}
	\vspace{-3pt}
	\textbf{Until} termination criterion is met (max. $\#$iterations)\\
	\textbf{output:} ${\bf A}$, ${\bf B}$, ${\bf C}$
\end{algorithm}

\begin{table*}[t]%[t]%[htbp]%[!htb]
	%		\begin{minipage}{.5\linewidth}
	\caption{\small Summary of datasets and their statistics.}
	\centering
	\vspace{-3mm}
	\label{table:data}
	\resizebox{0.67\textwidth}{!}{
		\begin{tabular}{c| c c c c c c}
			\hline
			%& \multicolumn{2}{|c}{\bf{RED ($\%$)}}\\
			%\cline{1-6}	
			{\bf Dataset} ($\tensorX$) &   {\bf Size} & {\bf Max} & {\bf Avg} & {\bf SD} & {\bf  \% (missing entries)} & {\bf  \% (zero entries)}\\
			\hline
			{\bf BAT}h Soap   &  $93 \times 50 \times 266$ &  52 &  0.79 & 1.34 &  44.73 & 33.37\\
			{\bf B}ottled {\bf J}ui{\bf C}es   &  $93 \times 50 \times 393$ & 12288  & 13.76  & 50.08 & 8.79 & 9.19\\
			{\bf CHE}eses  & $93 \times 50 \times 393$ &  18176 &  26.65 & 88.29 & 8.59 & 5.51\\
			{\bf COO}kies  &  $94 \times 50 \times 390$ &  14080 &  16.00 & 56.86 & 9.81 & 7.57\\
			{\bf CRA}ckers  & $94 \times 50 \times 382$ &  14080 &  8.21 & 29.61 & 14.21 & 7.57\\
			{\bf C}anned {\bf SO}up   & $93 \times 50 \times 379$ &  34494 & 40.46  & 133.42 & 8.64 & 4.54\\
			{\bf F}abric {\bf S}o{\bf F}teners   & $93 \times 50 \times 397$ &  7168 &  5.68 & 18.84 & 18.64 &  27.48\\
			{\bf GRO}oming   &  $93 \times 50 \times 272$ &  232 & 1.94  & 2.94 &  7.66 & 32.66 \\
			{\bf P}aper {\bf T}o{\bf W}els  &  $93 \times 50 \times 389$ & 19712  &  45.36 & 117.82 &  36.72 & 23.49\\
			{\bf S}oft {\bf DR}inks &  $93 \times 50 \times 391$ & 18944  & 48.81  & 155.09 &  8.58 & 11.18\\
			\hline
			{\bf Mixed DFF}   & $93 \times 500 \times 230$ &  17610 &   19.01 &  71.30 & 15.30 & 17.83\\
			\hline
			\hline
			{\bf Walmart} & $45 \times 99 \times 143$ &  6.93e+05 &  1.05e+04 & 1.99e+04 & 0 &    33.84\\
			\hline
			{\bf Crime} & $304 \times 388 \times 221$ &  325 & 0.26 & 1.47 & 0 &     91.56\\
			\hline
			{\bf Weather} & $49 \times 17 \times 365$ & 1038  & 10.23 & 95.65 & 0 & 93.30 \\
			\hline
	\end{tabular}} 
	\setlength{\textfloatsep}{0.1cm}
	\addtolength{\parskip}{-0.5mm}
	\vspace{-4mm}
\end{table*}

\vspace{3pt}
\noindent {\bf Walmart\footnote{{https://www.kaggle.com/c/walmart-recruiting-store-sales-forecasting/data}}:} Historical weekly sales data for $99$ different departments in $45$ Walmart stores located in different regions. A (stores $\times$ departments $\times$ weeks) tensor is created from these data. The resulting tensor is complete and has no missing entries. The size of each store (in square feet) is included in the data {(we use this information to form groups of stores)}.

\vspace{3pt}
\noindent {{\bf Crime\footnote{{https://www.kaggle.com/chicago/chicago-crime/activity}}:} Reported incidents of crimes that occurred in the city of Chicago from 2001 to present. Each incident is marked with its beat (police geographical area), and a code indicating the crime type. There are $304$ geographical areas and $388$ crime types in total. {Using this dataset, we form a (locations (by beat) $\times$ crime types $\times$ months) tensor.}} 

\vspace{3pt}
\noindent {\bf Weather\footnote{{http://www.bom.gov.au/climate/data/}}}: Daily weather observations from $49$ stations in Australia. These observations contain $17$ different variables, e.g., min temperature, max temperature, cloud, humidity, wind, etc. We form a (station (location) $\times$ variables $\times$ days) tensor using one year of daily observations.

Table \ref{table:data} summarizes the different datasets described above, with their size, maximum and average values, Standard Deviation (SD), and percentage of missing entries and zeros. These datasets are the ground truth in our experiments, and represented by $\tensorX \in\R^{I \times J \times K}$.

%\vspace{-5pt}
\subsection{Aggregation Configuration}
\label{sec:config}

The aggregated observations (compressed tensors), that are used as inputs to the disaggregation methods, are generated from $\tensorX$ following two practical scenarios described below: 

\vspace{3pt}
\noindent{\bf Scenario A}: The multidimensional data, we aim to disaggregate, are represented by $\tensorX \in\R^{I \times J \times K}$. 
Instead of the full tensor $\tensorX$, we are given two aggregated views: 
1)~temporally aggregated tensor $\tensorYt = \tensorX \times_3 {\bf W}$, i.e., aggregated in the third dimension; and 2) contemporaneously aggregated tensor $\tensorYc = \tensorX \times_1 {\bf U}$, aggregated in the first mode (e.g., stores/locations dimension).
We use the $10$ category-specific datasets from DFF and Walmart data to test this scenario. The stores are aggregated according to their geographical locations in the DFF datasets, and based on their sizes in Walmart data. 
We also test this scenario on Weather data, where the temporal aggregate represents the weather observations averaged over a course of time, and the contemporaneous aggregate is the average of the observations over a geographical region.

\vspace{3pt}
\noindent{\bf Scenario B}: In this scenario, 
two aggregated views of $\tensorX$ are given: 
1) similar to the previous scenario, temporally aggregated tensor $\tensorYt = \tensorX \times_3 {\bf W}$; and 
2) contemporaneously aggregated tensor $\tensorYc = \tensorX \times_1 {\bf U} \times_2 {\bf V}$, aggregated in the first \emph{and} second dimensions (e.g., sales counts that are \emph{jointly} aggregated over groups of stores {\em and} groups of items).
We use Mixed DFF and Crime data to test this scenario. 
The stores are aggregated into groups according to their locations in Mixed DFF data, whereas items are aggregated according to their categories.
In Crime data, locations and types are grouped based on the closeness in geographical location and similarity in crime type, respectively. 
Note that when ${\bf V} = {\bf I}$, this yields to Scenario A. 
Evidently, this scenario is more challenging since the second observation is aggregated in two modes, 
i.e., double aggregation, resulting in fewer measurements.

The difficulty of the problem {also} depends on the {\em aggregation level}, i.e., the number of data points (e.g., weeks, items, or stores) in one sum. Fewer aggregated measurements result in more challenging problems {from an ``equations versus unknowns” standpoint.
	We test the disaggregation performance using different aggregation levels for each dimension.}

%\vspace{-4pt}
\subsection{Evaluation Baselines \& Metrics}
\label{sec:metric}
We evaluate the disaggregation performance of the proposed method using 
the Normalized Disaggregation Error (NDE =  $\|\tensorX - \widehat{\tensorX}\|_F^2 / \|\tensorX\|_F^2$), where $\widehat{\tensorX}$ is the estimated data.  
The baseline methods are described next. Note that we compare to state-of-art approaches in the time series disaggregation literature as well as
methods developed to fuse multiple views of multidimensional data, but for different tasks. To the best of our knowledge our work is the first to perform disaggregation on multidimensional data from multiple views.

\vspace{3pt}
\noindent{\bf Mean:}
This baseline assumes that the constituent data atoms (entries in ${\bf \tensorX}$) have equal contribution in their aggregated samples. The final estimate of Mean is the average of the estimation from the temporal and the contemporaneous aggregates. 
{For example, the contemporaneous aggregate reports $100$ units sold in $10$ stores in the first week of January, and the temporal one tells us that $80$ units were sold in January (4 weeks) in Store~1. Then, Mean estimation of week~1 and store~1 is $(100/10 + 80/4)/2 = 15$ }

\vspace{3pt}
\noindent{\bf LS:}
This baseline is inspired by \cite{pavia2007estimating,di1990estimation}, where a least squares criterion is adopted on the linear relationship between the target time series in high resolution and the available aggregates. 
The resulting linear system is underdetermined, thus, these works assume a linear regression model between the target series and some set of indicators. In their context, indicators are time series available in high resolution that are expected to display similar fluctuations to the target series. For example, {the stock price of an oil company is a linear combination of the stock prices of other relevant companies.} 
This assumption requires additional data that are not available in our datasets. 
Therefore, we resort to the minimum-norm solution
{
 \begin{equation}\label{LSnn}
\begin{aligned}
\min_{{\bf \tensorX}}  \quad & \|\text{vec}({{\bf \Omega}_3^t}^T) \circledast \big(\text{vec}({{\bf Y}_3^t}^T) - \widetilde{{\bf W}}\text{vec}({{\bf X}_3}^T)\big)\|_2^2 \\
& + \|\text{vec}({\bf \Omega}_3^c) \circledast \big(\text{vec}({\bf Y}_3^c) - \widetilde{{\bf U}}\text{vec}({\bf X}_3)\big)\|_2^2 \\
\end{aligned}
\end{equation}
where $\widetilde{{\bf W}} = {\bf I} \otimes {\bf W}$ and $\widetilde{{\bf U}} = {\bf I} \otimes {\bf V} \otimes {\bf U}$.}

\vspace{3pt}
\noindent{\bf H-Fuse:}\cite{liu2017h}
This baseline constrains the solution to the LS baseline above to be smooth, i.e., it penalizes large differences between adjacent time ticks.

\vspace{3pt}
\noindent {\bf HomeRun:}\cite{almutairi2018homerun}
To circumvent the indeterminacy of the linear system in the time series disaggregation problem, this baseline
solves for the disaggregated series in the frequency domain. 
More specifically, HomeRun searches for the coefficients of the Discrete Cosine Transform (DCT) that represent the target high-resolution series. 
The key point is that the number of non-negligible DCT coefficients of the time series is much smaller than its length. 
In other words, the DCT is used as a sparsifying dictionary to reduce the number of variables.   
HomeRun also imposes smoothness and non-negativity constraints.

\vspace{3pt}
\noindent{\bf CMTF:} Couple Matricized Tensor Factorization has been widely used,
to fuse multiple views of multidimensional data,
in the hyperspectral imaging application \cite{yokoya2011coupled, simoes2014convex}---the work in \cite{yokoya2011coupled} adds non-negativity constraints.
These images are three-dimensional tensors, and the motivation behind these works is to exploit the low-rankness of the matricized image. We compare to this model because real world multidimensional data are often well-approximated using low-rank, as we will show empirically. Using our notation, CMTF solves 
\begin{equation}\label{eq:CMTF}
\begin{aligned}
\min_{{\bf A},{\bf B}} \quad & \|{\bf \Omega}_3^t \circledast ({\bf Y}^t_3 -  {\bf A}({\bf WB})^T)\|_F^2\\
& + \|{\bf \Omega}_3^c \circledast ({\bf Y}^c_3 -  ({\bf V} \otimes {\bf U}){\bf A}{\bf B}^T)\|_F^2 
\end{aligned}
\end{equation}
We solve \eqref{eq:CMTF} using a BCD algorithm with exact line search. Similar to \tensorglue, a good initialization for the low-rank factors improves the performance of CMTF. To ensure fair comparison, we initialize using SVD with missing entries set to be zeros.

In addition to the above baseline methods, we also test the estimation of the target disaggregated data with the following {\em oracle} baseline.

\vspace{3pt}
\noindent{\bf CPD}: We fit a CPD model directly to the ground truth tensor ${\tensorX}$ \emph{with respect to the observed entries}. We use the Matlab-based package \texttt{Tensorlab} to compute the CPD. Then, we reconstruct $\widehat{\tensorX}$ from the learned factors (${\bf A}, {\bf B}, {\bf C}$). This baseline can also serve as a lower bound for the error produced by the proposed method \tensorglue.

\begin{table*}[t]%[htbp]
	\caption{NDE of the proposed methods and the baselines using the $10$ category-specific datasets.}
	\centering
	\vspace{-1mm}
	\label{tab:nmsr}
	\resizebox{0.75\textwidth}{!}{
		\begin{tabular}{ c|| c | c |c | c | c | c | c | c | c | c }
			%\cline{2-11} 
			{\bf Dataset}	&  {\bf BAT}  & {\bf BJC}  & {\bf CHE}  & {\bf COO} & {\bf CRA} & 
			{\bf CSO} &  {\bf FSF} & {\bf GRO} & {\bf PTW} &  {\bf SDR}\\
			\hline
			\%missings	&  $44.73\%$  & $8.79\%$  & $8.59\%$  & $9.81\%$ & $14.21\%$ & 
			$8.64\%$ & $18.64\%$ & $7.66\%$  & $36.72\%$ & $8.58\%$\\
			
			SD	    & 1.34 & 50.08 &  88.29 & 56.86 & 29.61 & 
			133.42 & 18.84 & 2.94 & 117.82 & 155.09 \\
			\hline  
			\hline 
			{\bf Mean}     & {\bf 0.3284} & 0.4441 & 0.3118 & 0.3596 &  \underline{0.5217} & 
			0.3309 & 0.5609 & \underline{0.246}4 & {0.2994} & 0.2860 \\
			\hline
			{\bf LS}      & \underline{0.3328} & 0.6077 &  0.4650 & 0.6224 & 0.5889 & 
			0.4664 & 0.5982 & 0.2831 &  0.4593 & 0.5420 \\
			
			\hline
			{\bf H-FUSE}  & 0.3411 & 0.6437 &  0.4870 & 0.6414 & 0.5726 & 
			0.4885 & 0.6451 & 0.2863 &  0.4719 & 0.5644 \\
			\hline
			{\bf HomeRun} & 0.3461 & 0.6453 &  0.4818 & 0.6284 & 0.5376 & 
			0.4856 & 0.6496 & 0.2877 &  0.4662 & 0.5594 \\
			\hline
			{\bf CMTF}     & 0.4254 & \underline{0.1818} & \underline{0.1954} & \underline{0.1783} &  0.7455 & 
			\underline{0.1564} & \underline{0.1930} & 0.2908 & \underline{0.2577} &  \underline{0.1633} \\

			\hline
			{\bf \tensorglue}, R=10 & 0.5203 & 0.1978 & 0.1756 & 0.1757 &  {\bf 0.2587} & 
			0.2057 & 0.2019 & 0.3198 & 0.2844 &  0.2039 \\
			{\bf \tensorglue}, R=25 & 0.5079 & 0.1684 & 0.1516 & 0.1371 &  {0.2624} & 
			0.1373 & 0.1790 & 0.2581 & 0.2132 &  0.1438 \\
			{\bf \tensorglue}, R=40 & 0.4972 & {\bf 0.1572} & {\bf 0.1491} & {\bf 0.1318} &  0.2589 & 
			{\bf 0.1332} & {\bf 0.1747} & {\bf 0.2458} & {\bf 0.1969} & {\bf 0.1348}\\

			\hline
			{\bf CPD} (oracle), R=10          & 0.4782 &  0.0937 & 0.0723 & 0.1205 &  0.0776 & 
			0.0776 & 0.0810 & 0.2919 & 0.2356 &  0.1329 \\
			{\bf CPD} (oracle), R=25          & 0.4345 & 0.0586 & 0.0419 & 0.0676 &  0.0518 & 
			0.0476 & 0.0494 & 0.2448 & 0.1358 & 0.0822 \\
			{\bf CPD} (oracle), R=40          & 0.4109 & 0.0443 & 0.0321 & 0.0532 &  0.0438 & 
			0.0345 & 0.0399 & 0.2284 & 0.1007 & 0.0605 \\
			\hline
			\hline
			{\bf \blind}, R=10        & 0.5242 & 0.3012 & 0.3525 & 0.2207 &  0.3080 & 
			0.1752 & 0.2090 & 0.3156 & 0.3594 &  0.2008 \\
			{\bf \blind}, R=25        & 0.5002 & 0.3583 & 0.3553 & 0.2496 &  0.2976 & 
			0.1756 & 0.1892 & 0.2557 & 0.3758 &  0.1539 \\
			{\bf \blind}, R=40        & 0.4914 & 0.3909 & 0.3823 & 0.2942 &  0.3042 & 
			0.1825 & 0.1846 & 0.2472 & 0.3963 & 0.1620 \\                        
			\hline
	\end{tabular}}
	\vspace{-4mm}
\end{table*} 

\section{Experimental Results}
\label{sec:results}
\label{sec:effc}
In this section, we evaluate the performance of \tensorglue and \blind in terms of disaggregation accuracy using real data.
The two aforementioned aggregation scenarios (refer to Section \ref{sec:config}) are considered with different aggregation levels.
In the experiments, we choose the rank $R$ for \tensorglue (and the CPD baseline) based on Proposition \ref{thm:known}, unless stated otherwise.  
On the other hand, for CMTF, we perform a grid search and show the results with the best $R$.
{We run 10 iterations of the CPD step in the initialization of \tensorglue in Algorithm~\ref{alg:algo1} (or \blind in Algorithm \ref{alg:algo2}) using \texttt{Tensorlab}, then run 10 iterations of the iterative procedure in the algorithms.}
We set $\mu = 100$ for \blind in Algorithm~\ref{alg:algo2}. 
All experiments were performed using Matlab on a Linux server with an Intel Core i7--4790 CPU 3.60 GHz processor and 32 GB memory.

\subsection{Results on Scenario A} Two aggregated views $\tensorYt$, $\tensorYc$ are observed. 
Table \ref{tab:nmsr} shows the disaggregation error in terms of NDE, achieved by the proposed method and the baselines on the $10$ category-specific datasets from DFF. 
The proposed methods, \tensorglue and \blind, along with the CPD oracle are shown under 3 different ranks ($R=10,~R=25,~R=40$).
In $\tensorYt$, the weekly sales counts are observed on a monthly basis, while in $\tensorYc$, the $93$ (or $94$ for some categories) stores are clustered geographically into $18$ areas. 
This means that the measurements in the temporal aggregate $\tensorYt$ are about $25\%$ of the original size, and the number of the contemporaneously aggregated measurements in $\tensorYc$ is only $19.35\%$ of the disaggregated data size.

For all datasets in Table \ref{tab:nmsr}, except BAT, \tensorglue markedly outperforms the baselines---to highlight the improvement, we make the smallest error in bold and underline the second smallest. The naive mean (Mean) is good enough with BAT dataset because it is smooth (SD = $1.34$) and has the largest percentage of missing entries, compared to the other datasets. 
The time series methods, H-Fuse and HomeRun, do not perform well with these datasets because they are designed for smooth and quasi-periodic data, respectively. To provide an example, we noticed that HomeRun improves the error of LS and  H-Fuse baselines with CRA data, and found that CRA exhibit more periodicity compared to the rest of the categories. 
Comparing \tensorglue with CPD, we see that \tensorglue achieves error very close to CPD of the ground truth data with the same rank, e.g., with GRO, PTW, and SDR datasets. 
By looking at the performance of \blind in the table, we can see that the proposed algorithm works remarkably well when the aggregation matrices are unknown. For example, with GRO data and $R=40$, the NDE of \blind is $0.2472$, while NDE $=0.2284$ with CPD. \blind disaggregates with smaller, or very similar, error compared to the baselines that uses the aggregation pattern information---see results with CRA, FSF, GRO, and SDR datasets. 
With all datasets, there is always a wide range of $R$ under which the proposed algorithm works similarly~well.

\begin{figure}[t]%[htbp]%[t]
	\centering
	\subfloat[FSF dataset]
	{\includegraphics[width=0.25\textwidth,height=3.7cm]{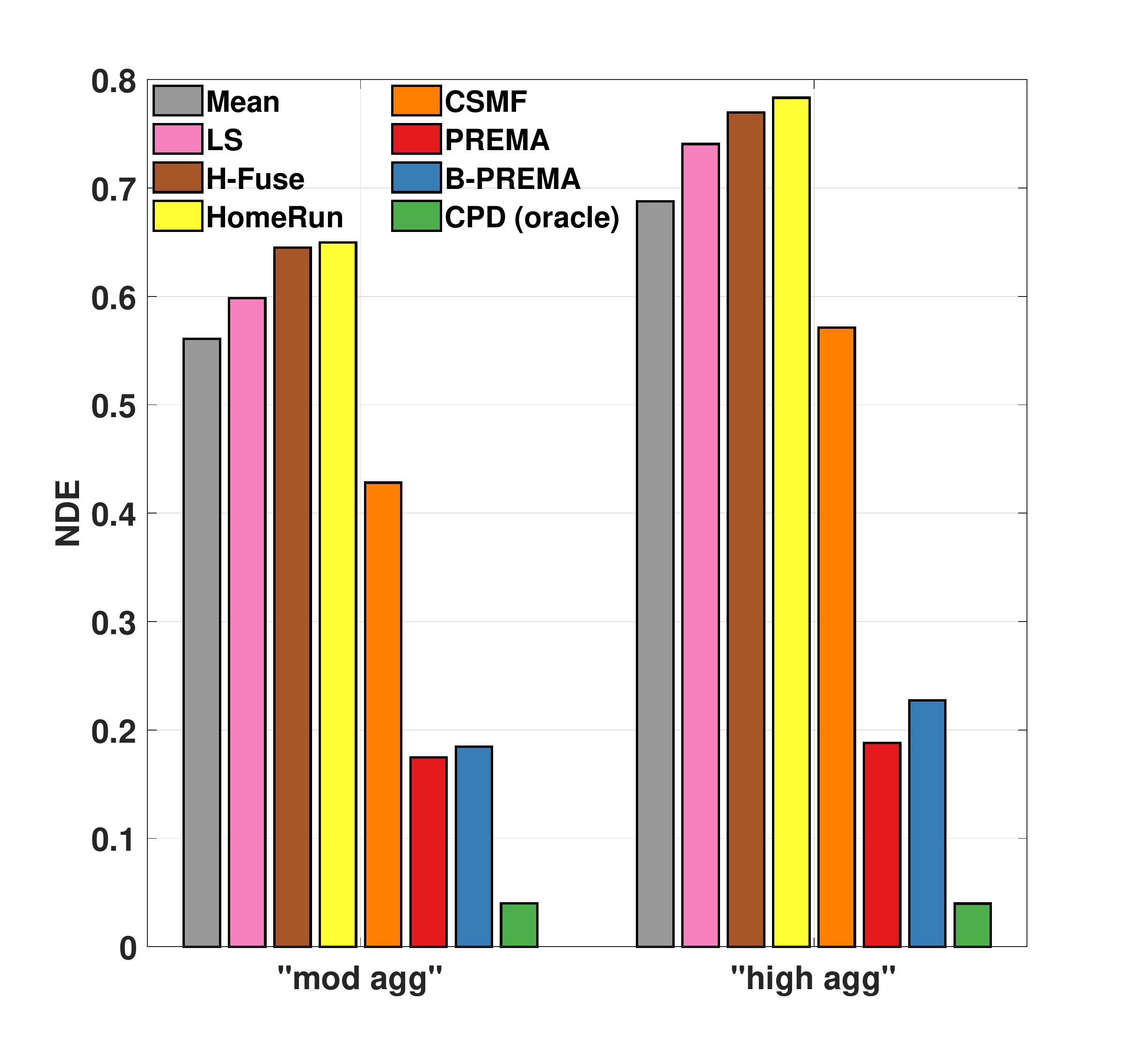}}\label{fig:R1a}
	\hspace{-13pt}
	\subfloat[PTW dataset]
	{\includegraphics[width=0.25\textwidth,height=3.7cm]{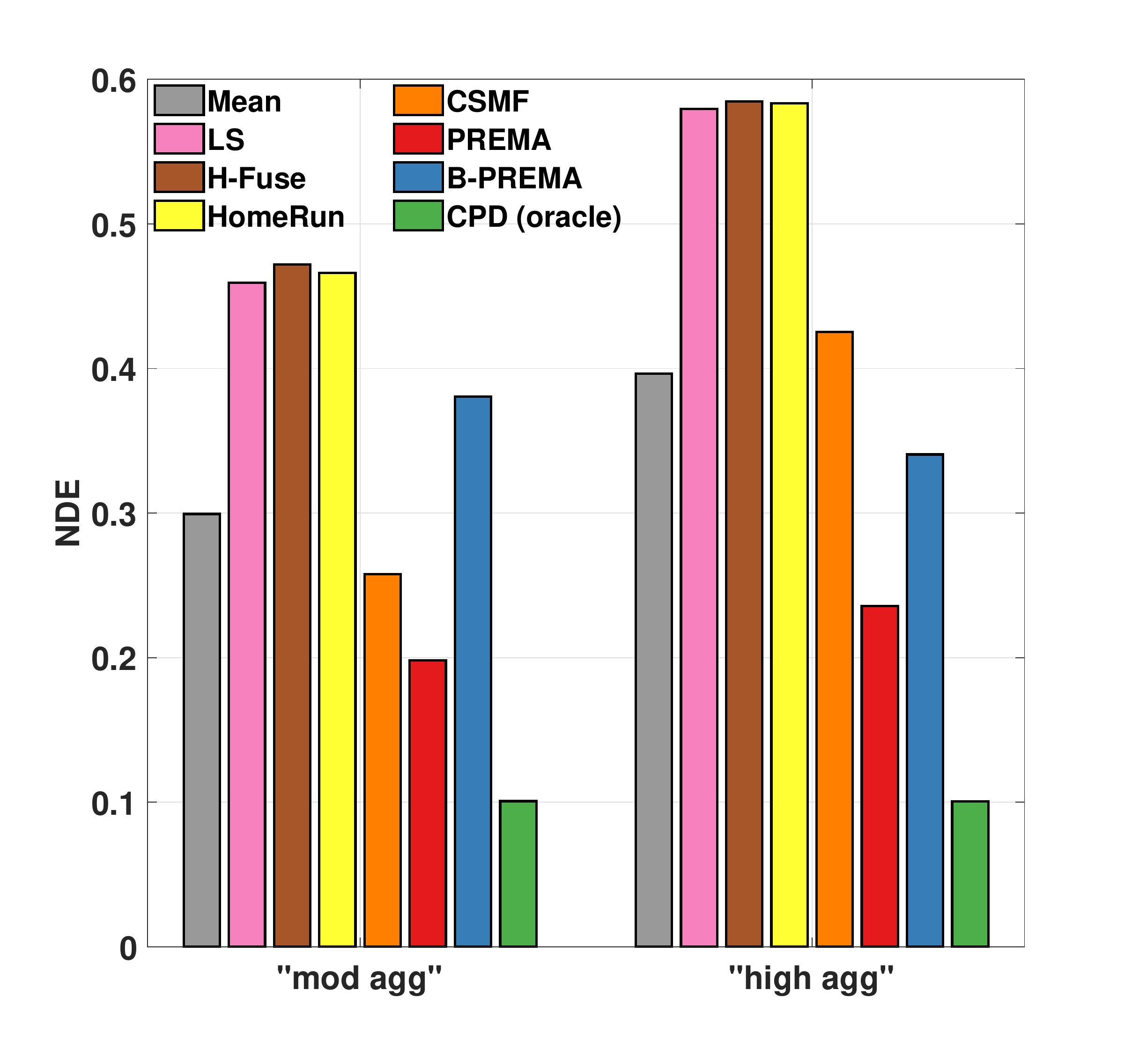}}\label{fig:R1b}
		\setlength{\belowcaptionskip}{-8pt}
		\vspace{-4pt}
	\caption{\tensorglue works well with extreme aggregation.}
	\label{fig:dff}
\end{figure}

\begin{figure}[t]%[htbp]%[t]
	\centering
	\subfloat[Walmart dataset]
	{\includegraphics[width=0.25\textwidth,height=3.8cm]{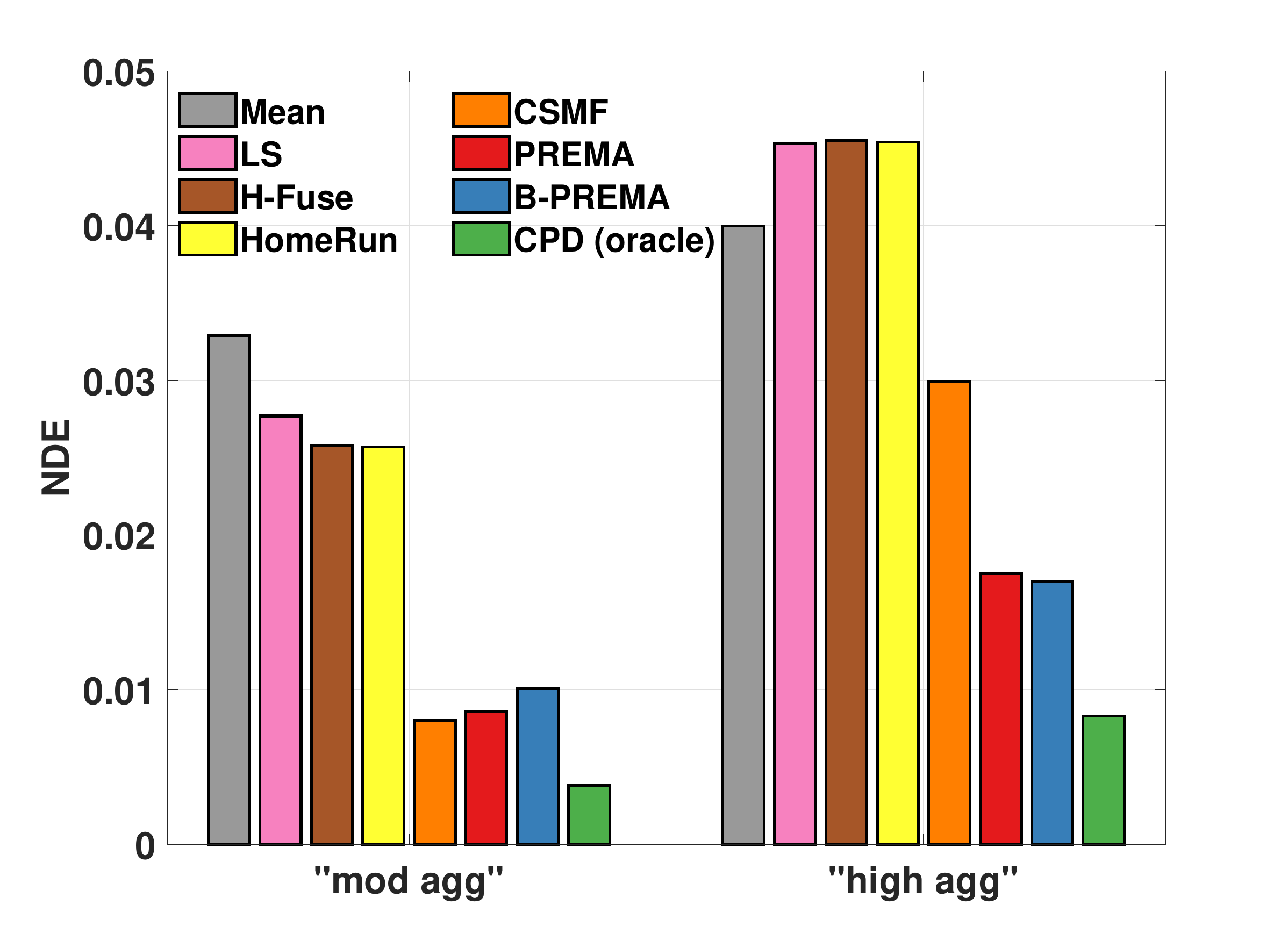}}\label{fig:R1a2}
	\hspace{-15pt}
	\subfloat[Weather dataset\footref{note1}]
	{\includegraphics[width=0.25\textwidth,height=3.8cm]{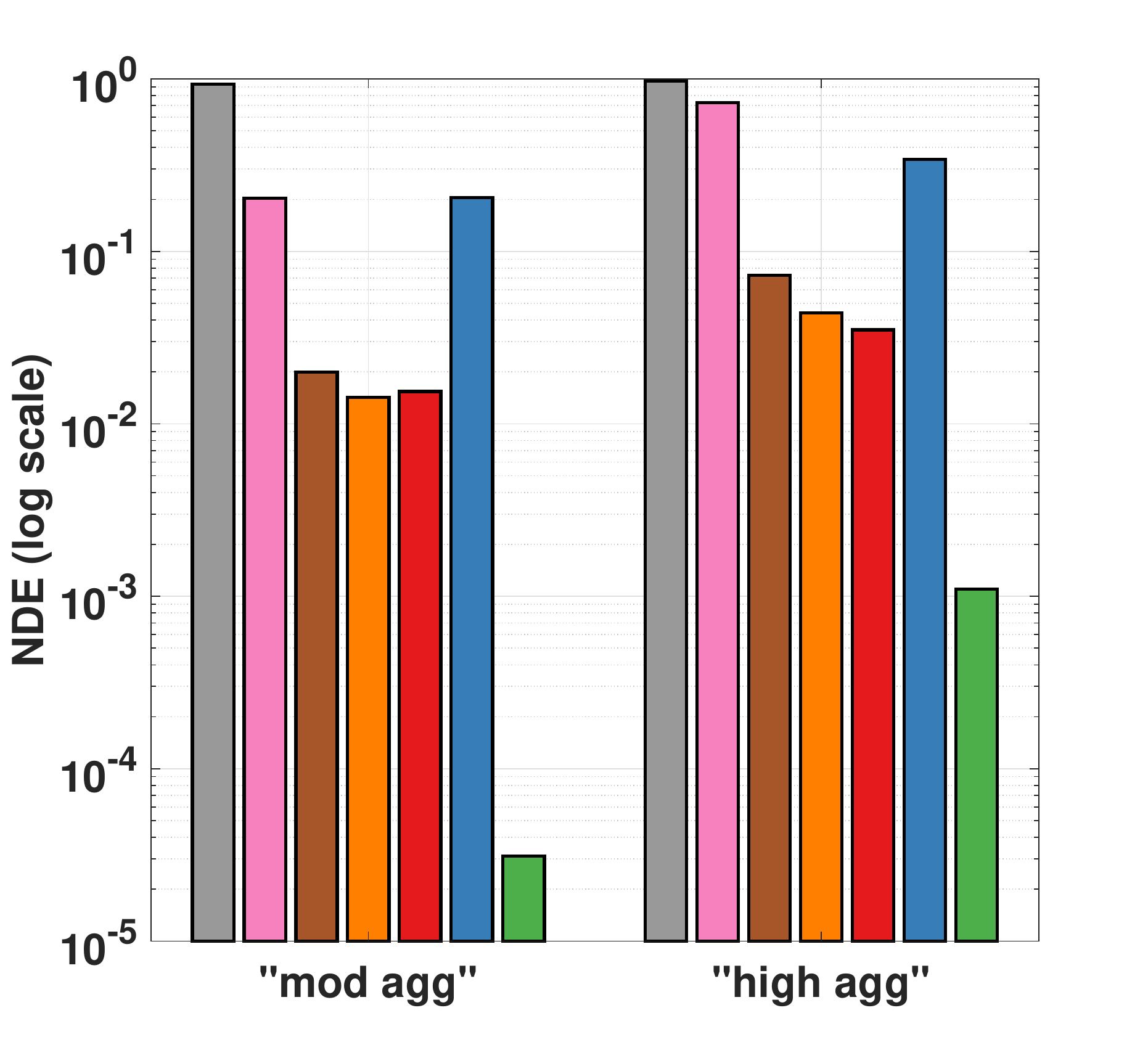}}\label{fig:R1b2}
		\setlength{\belowcaptionskip}{-18pt}
		\vspace{-4pt}
	\caption{\tensorglue works with different data.}
	\label{fig:ww}
\end{figure}

Next, we examine the performance when we change the level of aggregation from moderate (``mod agg") to very high (``high agg"). The disaggregation error is shown with two datasets from DFF data, FSF and PTW, in Figure~\ref{fig:dff}, and with Walmart and Weather\footnote{HomeRun is excluded from results with Weather data as it has non-negativity constraints.}  datasets in Figure~\ref{fig:ww}.  

The aggregation levels in Figure \ref{fig:dff} are: 1) monthly basis measurements (every 4 weeks) in $\tensorYt$, and the $93$ stores are divided geographically into $18$ areas (``mod agg"); and 2) quarterly samples (every 12 weeks) in $\tensorYt$, and the stores are divided into only $9$ areas (``high agg"). 
The rank $R$ for \tensorglue, \blind, and CPD is set to 40 in this figure.  
By comparing the moderate and high aggregation levels in Figure \ref{fig:dff}, we conclude that \tensorglue is more robust with aggressive aggregation where only few samples are available. 
With ``high agg", the number of aggregation samples is only $8.56\%$ of the original size in the temporal aggregate, and $9.68\%$ in the contemporaneous aggregate. In this case, the NDE of the best baseline is $3.04~(1.68)$x the error of \tensorglue with FSF (PTW) dataset, respectively.
PTW dataset is more challenging as it has relatively high percentage of missing entries ($36.72\%$).   
Moreover, with no knowledge of the aggregation pattern, \blind outperforms all baselines that have access to the aggregation information with FSF data. 
Although, \blind has NDE larger than Mean and CMTF with ``mod agg" on PTW data, it becomes superior to all baselines when the aggregation level is high.    

{With Walmart data in Figure \ref{fig:ww} (a), ``mod agg" means that weeks are aggregated into months in $\tensorYt$, and the $45$ stores are divided into $15$ groups, whereas time is aggregated quarterly (12 weeks) and stores are clustered into $9$ groups in ``high agg". CMTF works slightly better when the aggregation is moderate, owing to the fact that the second mode in Walmart data is departments as apposed to items in DFF data. Departments are less correlated than items from the same category. As a result, the advantage of tensor models over the matricized tensor in capturing the higher-order dependencies becomes less clear. However, \tensorglue is more immune to aggressive aggregation. In the ``high level" case, 
The NDE of CMTF is 1.71 times the error of \tensorglue. 
Even without access to the aggregation information, \blind significantly reduces the error of the baselines.}

In Figure \ref{fig:ww} (b), ``mod agg" corresponds to the daily weather measurements averaged into weekly samples, and the $49$ stations are averaged over $13$ stations. On the other hand, the daily measurements are averaged over monthly samples, and the $49$ stations are clustered into $7$ stations in the ``high agg" case.     
\tensorglue, CMTF, and H-Fuse perform similarly with Weather data\footnote{\label{note1}HomeRun is excluded from results with Weather data as it has non-negativity constraints.} (it has $93.30\%$ zeros) with moderate aggregation. 
The size of the second dimension of Weather data is small ($J = 17$), thus, the advantage of a tensor model over a matricized tensor model is less clear.
H-Fuse works well with this data as it penalizes the large jumps between the adjacent time ticks (i.e., days), and weather data are well suited for such constraint. 
Nevertheless, 
\tensorglue improves the error of CMTF and H-Fuse when the aggregation level is high.   
Although \blind does not work as well as with other data, it still has
smaller error than the simple baselines (Mean and LS), especially with aggressive aggregation. 

\subsection{Results on Scenario B}
\begin{figure}[t]%[htbp]%[t]
	\centering
	\subfloat[Mixed DFF dataset]
	{\includegraphics[width=0.25\textwidth,height=3.8cm]{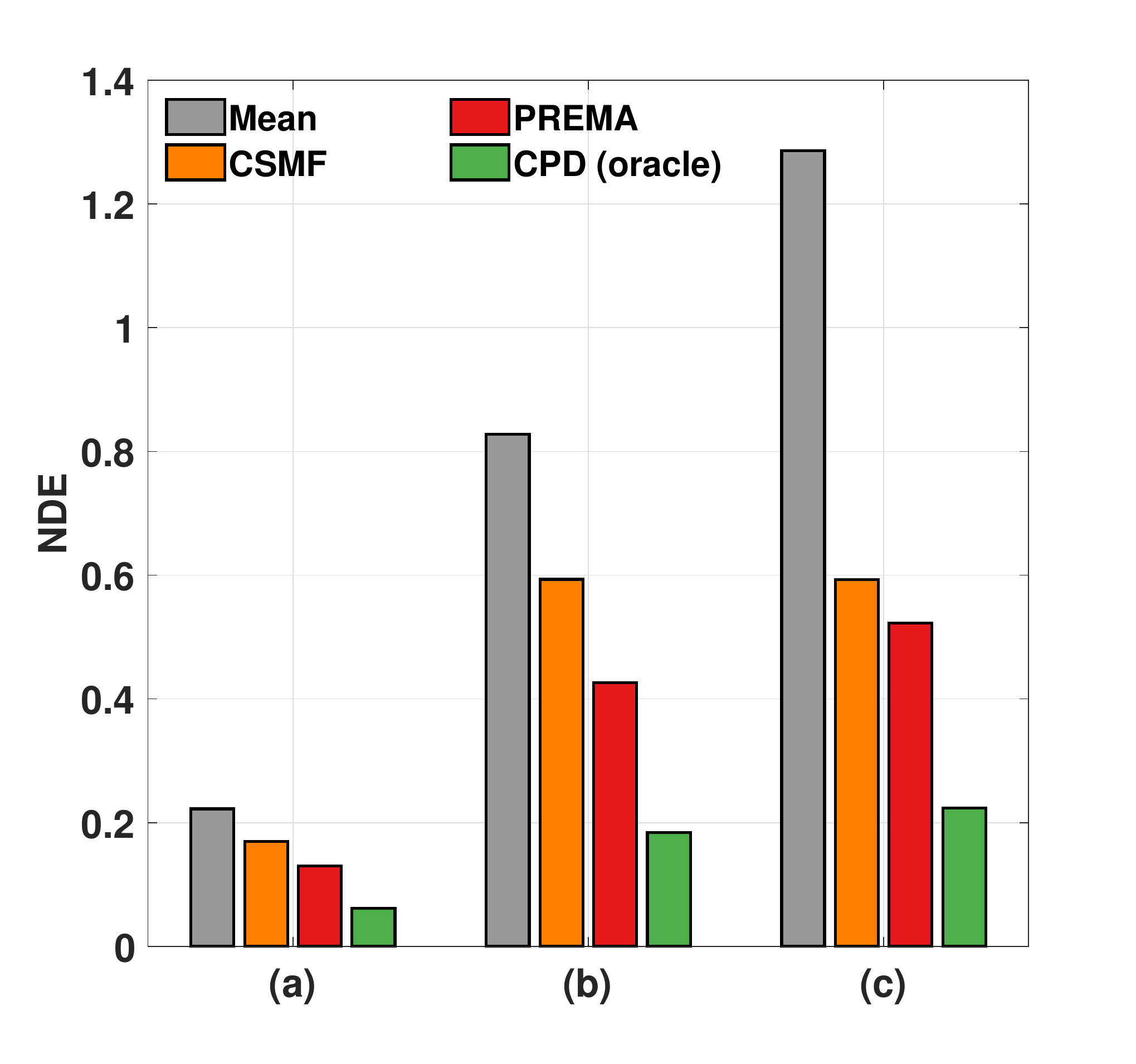}}\label{fig:R1a3}
	\hspace{-15pt}
	\subfloat[Crime dataset]
	{\includegraphics[width=0.25\textwidth,height=3.8cm]{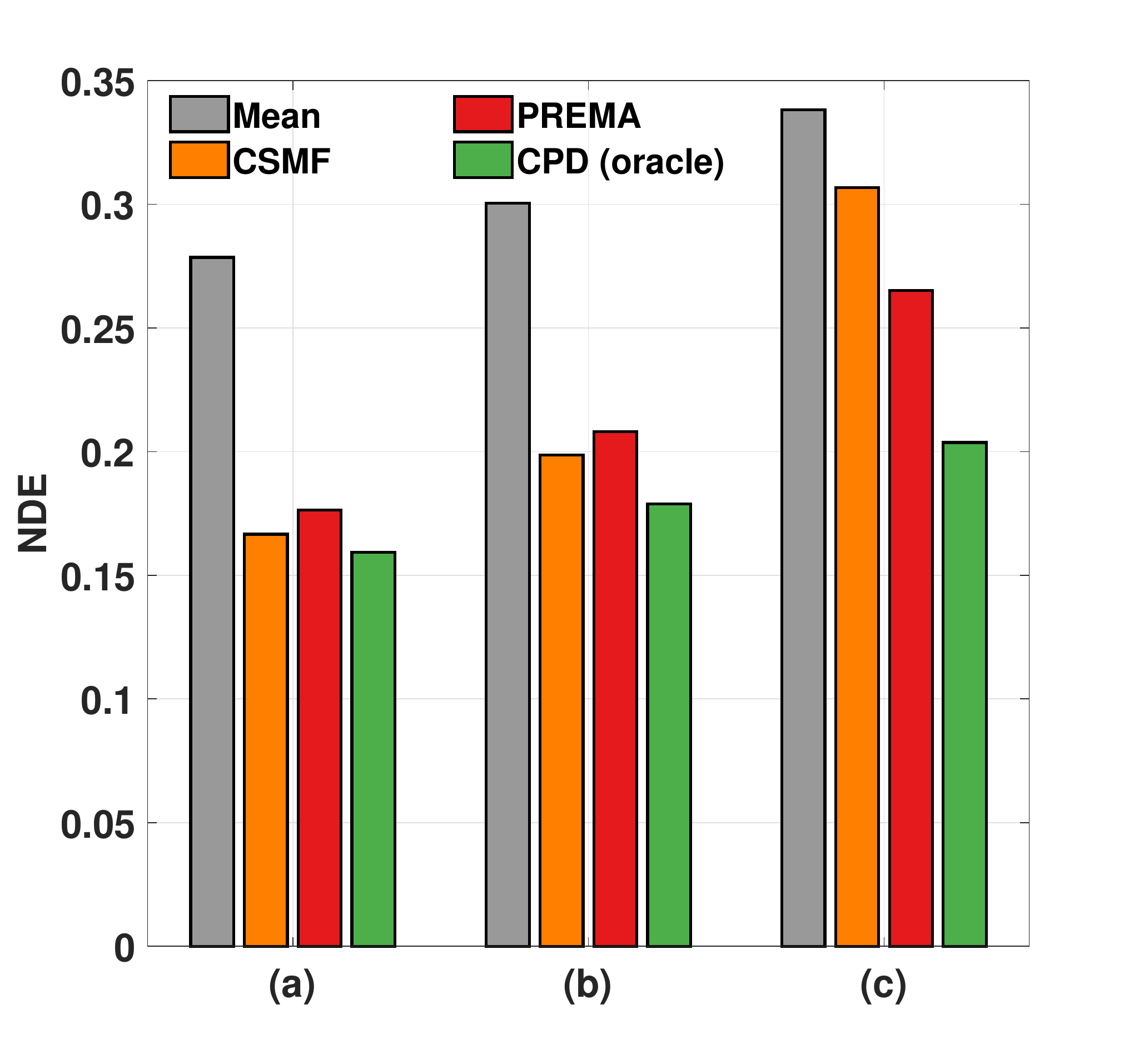}}\label{fig:R1b3}	\setlength{\belowcaptionskip}{-15pt}
	\vspace{-4pt}
	\caption{\tensorglue works with double aggregation (Scenario B).}
	\label{fig:crimemix}
\end{figure}
{The contemporaneous aggregate $\tensorYc$ in this scenario is aggregated in two dimensions: stores and items with Mixed DFF data, or crime locations and types with Crime data\footnote{LS,  H-Fuse,  and  HomeRun are excluded from this comparison as they run out of memory.}.  
We test this with three different aggregation levels with each data. Difficulty (i.e., level of aggregation), increases as we move from case (a) to (c)---Figure \ref{fig:crimemix} shows the performance for these three cases.}

{With Mixed DFF data, these levels are: a) $\tensorYt$ aggregates weeks into monthly samples, while $\tensorYc$ groups the $93$ stores into $18$ areas with no aggregation over the items, b) samples in $\tensorYt$ have monthly resolution, and $\tensorYc$ groups the stores into $18$ areas \emph{and} items into groups of $10$, and c) $\tensorYt$ contains temporal aggregates for each quarter of the year, and $\tensorYc$ groups stores into $18$ areas \emph{and} items into groups of $25$. 
One can see that the naive mean totally fails and its error exceeds $1$ in case~(c) with Mixed DFF data in Figure \ref{fig:crimemix}~(a).
Notwithstanding, \tensorglue works well with double aggregation {\em and} few available samples.}

{With Crime data, the aggregation levels are: a) $\tensorYt$ aggregates the months into quarterly resolution, while $\tensorYc$ clusters both the crime locations and types into groups of $5$, 
	b) $\tensorYt$ has a quarterly time resolution, and $\tensorYc$ aggregates both the locations and types into groups of $10$, and c) $\tensorYt$ aggregates the months into bi-yearly resolution, and $\tensorYc$ groups the crime locations and types into groups of $20$. Figure \ref{fig:crimemix}~(b) shows the performance with these levels using Crime data. These data are challenging as they have $91.56\%$ zero values and small SD.
	\tensorglue reduces the error of Mean significantly.  
	Although CMTF performs slightly better with the first two levels, \tensorglue becomes superior with extreme aggregation. }

\section{Conclusions}
\label{sec:con}
In this work, we proposed a novel framework, called \tensorglue, for fusing multiple aggregated views of multidimensional data. 
The proposed method leverages the properties of tensors in estimating the low-rank factors of the target data in higher resolution. The assumed model is provably transforming a highly ill-posed problem to an identifiable one.  
\tensorglue works with partially observed data, and can disaggregate effectively, even without any knowledge of the aggregation mechanism (\blind).  
Experimental results on real data show that the proposed algorithm is very effective, even in challenging scenarios, such as data with double aggregation and high level of  aggregation. 
The contributions of our work are summarized as  follows:
%\noindent
%\bit
%\vspace{1pt}
\begin{itemize}
	\item {\bf Formulation}: we formally defined the problem of multidimensional data disaggregation from views aggregated in different dimensions.  
	\item {\bf Identifiability:} The considered tensor model provably converts a highly ill-posed problem to an identifiable one. 
	\item {\bf Effectiveness:} \tensorglue reduced the disaggregation error of the competing alternatives by up to $67\%$.
	\item {\bf Unknown aggregation:} \blind works even when the aggregation mechanism is unknown. 
	\item {\bf Flexibility :} {\tensorglue can perform disaggregation on partially observed data.} 
	%\eit
\end{itemize}

%\end{comment}

%\begin{comment}

\appendices
\vspace{-2mm}
\section{Derivation of Gradient Expressions}
\label{sec:AppA}
%% %%%%%%%%%%%%%%%%%%%%%%%%%%%%%%%%%%%%%%%%%%%
{The terms in (11) and (24) can be divided into two types: 1)~CPD of a tensor, with some aggregation matrices multiplied with the factors; and 2) the regularization term $\mathcal{R}$ in (24). Because the gradient of a sum is the sum of the gradients, it is enough to show the derivation of the gradients using the function below. This function consists of two terms, each represents one of the terms types listed above. Consider: 
}
\begin{equation}\label{eq:general}
\begin{aligned}
\min_{{\bf A},{\bf B},{\bf C}} & ~\underbrace{\|\underline{{\bf \Omega}} \circledast (\tensorX -  ([\![{\bf U}{\bf A}, {\bf V}{\bf B},{\bf W}{\bf C}]\!])\|_F^2}_{\mathcal{T}}
+  \underbrace{\|{\bf 1}^T{\bf C} - {\bf 1}^T{\bf \widetilde{C}}\|_2^2}_{\mathcal{R}}
\end{aligned}
\end{equation}
where ${\bf \tensorW}$ is as defined in (10), and ${\bf \tensorX} \in \R^{I \times J \times K}$ is our data tensor. 
Note that all the CPD terms in (11) and (24) are similar to the term $\mathcal{T}$, with one (or more) of the aggregation matrices $\{{\bf U}, {\bf V}, {\bf W}\}$ is equal to ${\bf I}$. Thus, the term $\mathcal{T}$ generalizes all the CPD terms in our models. 
Using mode-3 unfolding, $\mathcal{T}$ is equivalent to
\begin{equation}\label{eq:genmod1}
\begin{aligned}
\mathcal{T} = \| {{\bf \Omega}}_3 \circledast  \big({\bf X}_3 -  (({\bf V}{\bf B}) \odot ({\bf U}{\bf A}))({\bf W}{\bf C})^T\big)\|_F^2
\end{aligned}
\end{equation}
Vectorizing the above, we get
\begin{equation}\label{eq:sy}
\begin{aligned}
\mathcal{T} = \|{\bf S}{\bf x} -  {\bf S}(({\bf V}{\bf B}) \odot ({\bf U}{\bf A})  \odot ({\bf W}{\bf C})){\bf 1}\|_F^2
\end{aligned}
\end{equation}
where ${\bf x} = \text{vec}({\bf X}_3)$, and {${\bf S} \in \{0,1\}^{N \times IJK}$ is a fat matrix with one 1 in each row to select the available entries in ${\bf x}$, where $N = nnz(\tensorW)$. Thus, the role of ${\bf S}$ with ${\bf x}$, is similar to the role of ${\tensorW}$ with the tensor form ${\tensorX}$.}
Equation \eqref{eq:sy} is also equivalent to  
\begin{equation}\label{eq:}
\begin{aligned}
\mathcal{T} = \|{\bf S}{\bf x} -  {\bf S} \big({\bf I} \otimes (({\bf V}{\bf B}) \odot ({\bf U}{\bf A}))\big) ({\bf W}  \otimes {\bf I}){\bf c}\|_F^2
\end{aligned}
\end{equation}
where ${\bf c}=$ vec(${\bf C}$). We show the derivative of $\mathcal{T}$ and $\mathcal{R}$ w.r.t. ${\bf C}$ (derivatives w.r.t. ${\bf A}$ and ${\bf B}$ are derived similarly by using mode-1 and mode-2 unfolding and rotating the factors accordingly).
{First, we derive the gradient of $\mathcal{T}$ w.r.t. ${\bf C}$:}
\begin{align}\label{eq:updatA_apdx}\nonumber
\nabla_{\bf C}\mathcal{T} & = 2({\bf W}^T \hspace{-1mm} \otimes {\bf I}) ({\bf I} \otimes ({\bf V}{\bf B} \odot {\bf U}{\bf A})^T){\bf S}^T{\bf S}\ ({\bf I} \otimes ({\bf V}{\bf B} \odot\cdot\\\nonumber
&{\bf U}{\bf A})) ({\bf W} \otimes {\bf I}){\bf c} - 2({\bf W}^T \hspace{-2mm} \otimes {\bf I})({\bf I} \otimes ({\bf V}{\bf B} \odot {\bf U}{\bf A})^T){\bf S}^T{\bf S}{\bf x}\nonumber\\\nonumber
&= 2({\bf I} \otimes ({\bf V}{\bf B} \odot {\bf U}{\bf A})^T)({\bf W}^T \otimes {\bf I}){\bf S}^T{\bf S}({\bf I} \otimes ({\bf V}{\bf B} \odot\cdot\\\nonumber
&{\bf U}{\bf A})) ({\bf W} \otimes {\bf I}){\bf c} - 2({\bf I} \otimes ({\bf V}{\bf B} \odot {\bf U}{\bf A})^T)({\bf W}^T\hspace{-2mm} \otimes {\bf I}){\bf S}^T{\bf S}{\bf x} \nonumber\\\nonumber
& = 2({\bf I} \otimes ({\bf V}{\bf B} \odot {\bf U}{\bf A})^T)({\bf W}^T \otimes {\bf I}){\bf S}^T{\bf S}  \big(({\bf I} \otimes ({\bf V}{\bf B} \odot \cdot\\
& {\bf U}{\bf A})) ({\bf W} \otimes {\bf I}){\bf c} - {\bf x}\big)
\end{align}
{We can use mode-3 unfolding to rewrite the final equation in \eqref{eq:updatA_apdx} above as
\begin{equation}\label{eq:updatAApdx}
\begin{aligned}
\nabla_{\bf A}\mathcal{T} & = 2{\bf W}^T\big({\bf \Omega}_3 \circledast ({\bf \widehat{X}}_3 - {\bf X}_3)\big)^T\big(({\bf VB}) \odot ({\bf UA})\big)\\
\end{aligned}
\end{equation}
where ${\bf \widehat{X}}_3 = \big(({\bf {VB}}) \odot ({\bf UA})\big)({\bf WC})^T$.}
{The gradient above can be computed efficiently by the following steps:
\vspace{-2mm}
\begin{enumerate}
    \item Compute ${\bf L} = {\bf \Omega}_3 \circledast ({\bf \widehat{X}}_3 - {\bf X}_3)$.
    \item Compute ${\bf M} = {\bf L}^T\big(({\bf VB}) \odot ({\bf UA})\big)$.
    \item Compute $2{\bf W}^T {\bf M}$
\end{enumerate}
}
\noindent 
{Next,} the derivative of $\mathcal{R}$ w.r.t. ${\bf C}$ is
\begin{equation}\label{eq:updatAApdxR}
\begin{aligned}
\nabla_{\bf C}\mathcal{R} & = 2({\bf 1}{\bf 1}^T{\bf C} 
- {\bf 1}{\bf 1}^T{\bf \widetilde{C}})
\end{aligned}
\end{equation}

%% %%%%%%%%%%%%%%%%%%%%%%%%%%%%%%%%%%%%%%%
%\vspace{-3.5mm}
\section{Derivation of Step Size Expressions}
\label{sec:AppB}
The step size terms in both Algorithm 1 and 2
are chosen using the exact line search optimization {method}.
{As mentioned earlier, the function \eqref{eq:general} generalizes all the terms in \tensorglue and \blind models. Thus, we use \eqref{eq:general} to show how to find the step size $\gamma$ associated with updating ${\bf C}$ as an illustrative example.
In this case, the exact line search chooses $\gamma$ to be the minimizer of
\begin{equation}\label{eq:}
\begin{aligned}
\argmin_{\gamma\geq 0} \quad & \mathcal{F}\big({\bf C} - \gamma\nabla_{\bf C}\mathcal{F}\big)
\end{aligned}
\end{equation}
where $\mathcal{F} = \mathcal{L} + \mathcal{R}$, which are as defined in \eqref{eq:general}. 
Plugging the variable ${\bf C} - \gamma\nabla_{\bf C}\mathcal{F}$ into \eqref{eq:general} and rearranging the terms, we get
\begin{equation}\label{eq:55}
\begin{aligned}
\argmin_{\gamma\geq 0} \quad &  \|  \underbrace{{{\bf \Omega}}_3 \circledast\big({\bf Y}_3 -  \big(({\bf V}{\bf B}) \odot ({\bf U}{\bf A})\big){\bf W}^T{\bf C}^T\big)}_{{\bf E}}\\
& + \gamma\underbrace{{{\bf \Omega}}_3 \circledast \big(({\bf V}{\bf B} \odot {\bf U}{\bf A}){\bf W}^T\nabla_{\bf C}\mathcal{F}^T\big)}_{{\bf D}}\|_F^2\\
& + \|\underbrace{{\bf 1}^T{\bf C} - {\bf 1}^T{\bf \widetilde{C}}}_{{\bf e}^T}
- \gamma \underbrace{{\bf 1}^T\nabla_{\bf C}\mathcal{F}}_{{\bf d}^T}\|_2^2
\end{aligned}
\end{equation}}
{One can see that at the optimal solution to \eqref{eq:55}, we have:
\vspace{-2mm}
\begin{align}
    -\text{vec}({\bf E})^T &= \gamma \text{vec}({\bf D})^T \label{aa}\\
    {\bf e}^T &= \gamma {\bf d}^T \label{bb}
\end{align}
Multiplying \eqref{aa} by vec(${\bf D}$) and \eqref{bb} by ${\bf d}$, and summing up the resulting two equations, we get
\begin{align}\label{aabb}
    - \text{vec}({\bf E})^T\text{vec}({\bf D}) +  {\bf e}^T{\bf d} 
    = \gamma (\text{vec}({\bf D})^T\text{vec}({\bf D}) + {\bf d}^T{\bf d})
\end{align}
}

\noindent {Respecting the non-negativity constraint, we can see that the optimal solution is 
\begin{equation}\label{eq:}
\gamma  = max\Big(0, \frac{ -\text{vec}({\bf E})^T\text{vec}({\bf D}) + {\bf e}^T{\bf d}}{\text{vec}({\bf D})^T\text{vec}({\bf D}) + {\bf d}^T{\bf d}}\Big),
\end{equation}}

\section{Initialization Algorithm}
\label{sec:AppC}
The initialization steps of Algorithm 1 are as follows

\vspace{3pt}
\begin{algorithmic}
	\STATE Set missing entries in $\tensorYt$, and $\tensorYc$ to zeros.
	\IF{${\bf V} = {\bf I}$ and $K>I$} 
	\STATE ${\bf \widetilde{A}}, {\bf B}, {\bf {C}} \leftarrow$
	{\fontfamily{qcr}\selectfont CPD}($\tensorYc$);
	\vspace{2pt}
	\STATE ${\bf {A}} \leftarrow \text{solve~} 
	{\bf Y}^t_1 = (({\bf {WC}}) \odot {\bf B}) {\bf A}^T$
	\ELSE
	\STATE ${\bf {A}}, {\bf B}, {\bf \widetilde{C}} \leftarrow$
	{\fontfamily{qcr}\selectfont CPD}($\tensorYt$);
	\vspace{2pt}
	\STATE ${\bf {C}} \leftarrow \text{solve~} {\bf Y}^c_3 = (({\bf {VB}}) \odot ({\bf {UA}})){\bf {C}}^T$  
	\ENDIF
\end{algorithmic}
Note that the missing entries are set to $0$ only in the initialization steps.
We use the Matlab-based package 
{\fontfamily{qcr}\selectfont Tensorlab}
to compute the CPD in the initialization (e.g., {\fontfamily{qcr}\selectfont CPD}($\tensorYc$)).

%\vspace{-3mm}
\section{Proof of Proposition 1}
\label{sec:AppD}
Let $\tensorX \in\R^{I \times J \times K}$ be the target tensor to disaggregate with CPD $\tensorX = [\![{\bf A}, {\bf B},{\bf C}]\!]$ of rank $R$ and $\tensorYt\in\R^{I \times J \times K_w} = \tensorX \times_3 {\bf W}$. Then under the conditions of Proposition 1, $\tensorYt$ admits a unique CPD $\tensorYt = [\![{\bf A}_t, {\bf B}_t,{\bf C}_t]\!]$. Since it is unique it holds that: \\
\begin{equation}
{\bf A}_t= {\bf A} {\bf\Pi}{\bf \Lambda}_1, {\bf B}_t={\bf B} {\bf\Pi}{\bf \Lambda}_2, {\bf C}_t={\bf W}{\bf C} {\bf\Pi}{\bf \Lambda}_3,
\end{equation}
where $\bf \Pi$ is a permutation matrix and ${\bf\Lambda}_1,~{\bf\Lambda}_2,~,{\bf\Lambda}_3$ are diagonal matrices such that ${\bf\Lambda}_1{\bf\Lambda}_2{\bf \Lambda}_3=\bf I$. In the case where $\tensorYt$ has missing entries the conditions under which $[\![{\bf A}_t, {\bf B}_t,{\bf C}_t]\!]$ are identifiable are stricter and depend on the pattern of misses. We can use the conditions in \cite{sorensen2019fiber,kanatsoulis2019tensor,ashraphijuo2017fundamental} for fiber, regular and random sampling respectively. So far factors $\bf A,~\bf B$ have been identified up to column permutation and scaling. What remains to be proven is that:
\begin{equation}\label{appeq}
\tensorWc \circledast\tensorYc= \tensorWc \circledast(\tensorX \times_1 {\bf U}\times_2 {\bf V}) =\tensorWc \circledast( [\![{\bf U}{\bf A}, {\bf V}{\bf B},{\bf C}]\!]) 
\end{equation}
yields a solution for ${\bf C}_c$ such that ${\bf C}_c={\bf C} {\bf\Pi}{\bf \Lambda}_3$. Equation \eqref{appeq} can be equivalently written as:
\begin{equation}\label{eq:c}
{\bf \Omega}_c{\bf y}_c={\bf \Omega}_c( \bf C\odot{\bf V}{\bf B}\odot{\bf U}{\bf A})\bf 1={\bf \Omega}_c( \bf I\otimes({\bf V}{\bf B}\odot{\bf U}{\bf A}))\bf c,
\end{equation}
where ${\bf y}_c, \bf c$ are vectorized versions of $\tensorYc,~{\bf C}^T$ respectively and ${\bf \Omega}_c\in\{0,1\}^{L_c\times I_uJ_vK}$ is a fat selection matrix, that selects the available entries of ${\bf y}_c$.  

Now let $\widetilde{\bf A}={\bf U}{\bf A}$ and $\widetilde{\bf B}={\bf V}{\bf B}$. Following \cite[Lemma 1]{kanatsoulis2018hyperspectral} $\widetilde{\bf A},~\widetilde{\bf B}$ are drawn from absolutely continuous non-singular distributions. Also let $\bf P =\widetilde{\bf B}\odot\widetilde{\bf A}$. Since $I_uJ_v\geq R$ the determinant of any $R\times R$ submatrix of $\bf P$ is a non-trivial analytic function of $\widetilde{\bf A},~\widetilde{\bf B}$. Therefore any $R\times R$ minor of $\bf P$ is non-zero almost surely \cite[Lemma 3]{gunning2009analytic} and any $R$ rows of $\bf P$ are independent.

Taking a closer look at matrix $\bf G=\bf I\otimes({\bf V}{\bf B}\odot{\bf U}{\bf A})=\bf I\otimes(\widetilde{\bf B}\odot\widetilde{\bf A})$ we observe that it is an $I_uJ_vK\times KR$ block diagonal matrix of the form:
\begin{equation}
\bf G=\begin{bmatrix}
\bf P &\bf{0}&\dots&\bf{0}\\
\bf{0}&\bf P&\dots&\bf{0}\\
\vdots  & \vdots & \ddots & \vdots\\
\bf{0}&\bf{0}&\dots&\bf P\\
\end{bmatrix}=\begin{bmatrix}
{\bf G}_1\\
{\bf G}_2\\
\vdots \\
{\bf G}_K\\
\end{bmatrix}
\end{equation}

Each block ${\bf G}_k$ corresponds to the $k-$th frontal slab of $\tensorYc$ and the rows between different ${\bf G}_k$'s are independent by construction. Since we have assumed that the minimum number of observed entries for each frontal slab is greater than or equal to $R$, then ${\bf \Omega}_c\bf G$ has full column rank equal to $KR$ and the solution for $\bf c$ in \eqref{eq:c} is unique with probability 1.
Plugging ${\bf A}_t,~{\bf B}_t$ in equation \eqref{eq:c} we get:
\begin{align}\label{eq:c}
{\bf \Omega}_c{\bf y}_c &= {\bf \Omega}_c({\bf C}\odot{\bf V}{\bf B}_t\odot{\bf U}{\bf A}_t){\bf 1}\\ &= {\bf \Omega}_c ({\bf C}\odot{\bf V}{\bf B}{\bf \Pi}{\bf \Lambda}_2\odot{\bf U}{\bf A}{\bf \Pi}{\bf \Lambda}_1)\bf 1\label{eq:c3}
\end{align}
Then the unique solution for ${\bf C}$ satisfies ${\bf C}_c={\bf C}{\bf \Pi}{\bf \Lambda}_3$ and $\widehat{\tensorX}=[\![{\bf A}_t, {\bf B}_t,{\bf C}_c]\!]$ disaggregates $\tensorYt,~\tensorYc$ to $\tensorX$ almost surely.

\ifCLASSOPTIONcaptionsoff
  \newpage
\fi



%\bibliographystyle{IEEEtran}


%\newpage
\bibliographystyle{IEEEtran}
\bibliography{references_Prema}


%\vfill


%\begin{comment}

\vspace{-10mm}
\begin{IEEEbiography}[{\includegraphics[width=1in,height=1.25in,clip,keepaspectratio]{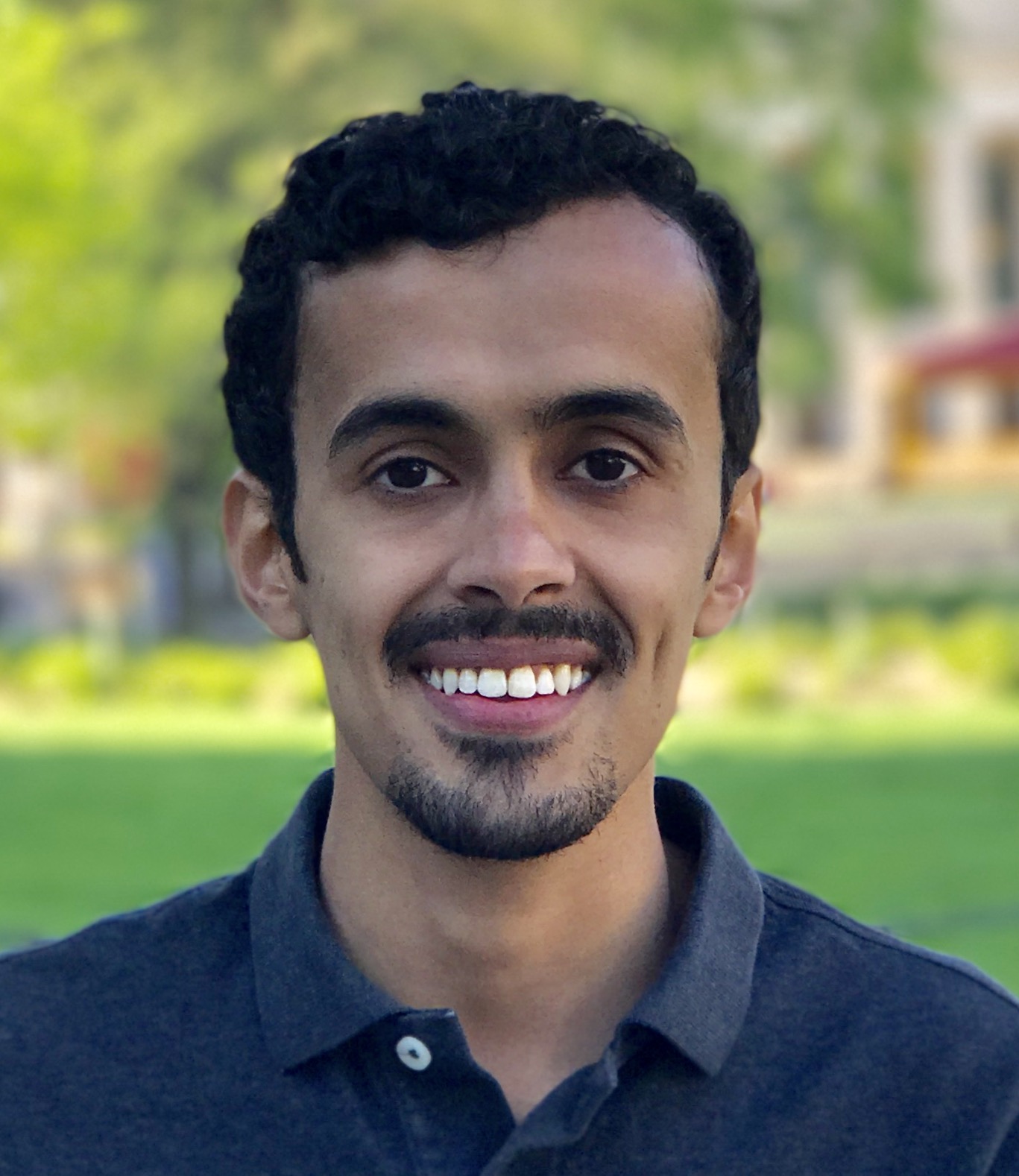}}]{Faisal~M.~Almutairi} received the B.Sc degree with first-class honor in electrical engineering from Qassim University (QU), Qassim, Saudi Arabia, in 2012. He received the M.S. degree in electrical engineering and the M.S. degree in Industrial Engineering from the University of Minnesota (UMN), Minneapolis, MN, USA, in 2016 and 2017, respectively. Faisal is currently a Ph.D. candidate in the Department of Electrical and Computer Engineering (ECE) at the UMN, Minneapolis, MN, USA. 
His research interests include signal processing, data mining, and optimization. 
\end{IEEEbiography}

\vspace{-10mm}
\begin{IEEEbiography}[{\includegraphics[width=1in,height=1.25in,clip,keepaspectratio]{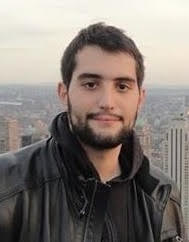}}]{Charilaos~I.~Kanatsoulis} (S'18) is a Ph.D. candidate in the Department of Electrical and Computer Engineering (ECE) at the University of Minnesota
(UMN), Twin Cities. He received his Diploma in electrical and computer engineering from the National Technical University of Athens, Greece, in 2014.  His research interests include signal processing, machine learning, tensor analysis, and graph mining.
\end{IEEEbiography}

%\vspace{-12mm}
\begin{IEEEbiography}[{\includegraphics[width=1in,height=1.25in,clip,keepaspectratio]{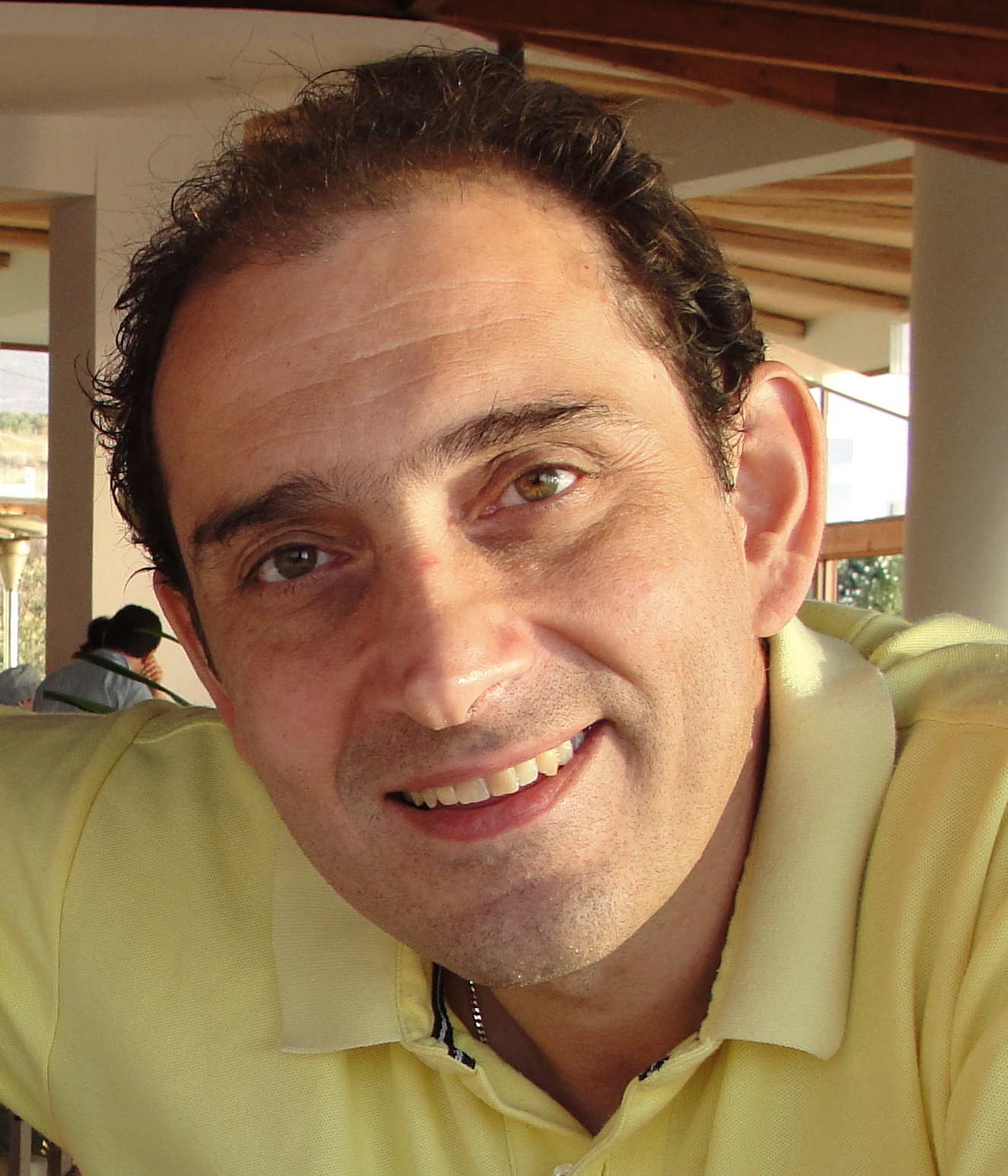}}]{Nicholas D. Sidiropoulos} 
(F’09) received the Diploma degree in electrical engineering from the Aristotle University of Thessaloniki, Thessaloniki, Greece, in 1988, and the M.S. and Ph.D. degrees in electrical engineering from the University of Maryland, College Park, MD, USA, in 1990 and 1992, respectively. He has served on the faculty of the University of Virginia (UVA), University of Minnesota, and the Technical University of Crete, Greece, prior to his current appointment as Louis T. Rader Professor and Chair of ECE at UVA. From 2015 to 2017, he was an ADC Chair Professor with the University of Minnesota. His research interests are in signal processing, communications, optimization, tensor decomposition, and factor analysis, with applications in machine learning and communications. He was the recipient of the NSF/CAREER award in 1998 and the IEEE Signal Processing Society (SPS) Best Paper Award in 2001, 2007, and 2011. He served as IEEE SPS Distinguished Lecturer during in 2008 and 2009, and as IEEE SPS Vice-President from 2017 to 2019. He was the recipient of the 2010 IEEE SPS Meritorious Service Award, and the 2013 Distinguished Alumni Award from the University of Maryland, Department of ECE. He became a Fellow of EURASIP in 2014.
\end{IEEEbiography}

%\end{comment}

\end{document}